\documentclass[10pt,twocolumn,letterpaper]{article}
\pdfoutput=1

\usepackage{iccv}
\usepackage{times}
\usepackage{epsfig}
\usepackage{graphicx}
\usepackage{amsmath}
\usepackage{amssymb}
\usepackage{nicefrac}

\usepackage{booktabs}
\usepackage{multirow}
\usepackage{xcolor}
\usepackage{caption}
\usepackage{subcaption}
\usepackage{array}
\usepackage{verbatim}
\usepackage{mathtools}
\usepackage{authblk}
\usepackage[accsupp]{axessibility}

\usepackage[breaklinks=true,bookmarks=false]{hyperref}

\iccvfinalcopy 

\def\iccvPaperID{17} 

\ificcvfinal\fi

\begin{document}

\title{A Hierarchical Assessment of Adversarial Severity}

\author[1]{Guillaume Jeanneret}
\author[2]{Juan C. Pérez}
\author[1]{Pablo Arbelaez}

\affil[1]{Center for Research and Formation in Artificial Intelligence, Universidad de los Andes}
\affil[2]{King Abdullah University of Science and Technology (KAUST)}
\affil[]{{\tt\small \{g.jeanneret10, pa.arbelaez\}@uniandes.edu.co}, $^2${\tt\small juan.perezsantamaria@kaust.edu.sa}}

\maketitle

\ificcvfinal\thispagestyle{empty}\fi

\begin{abstract}
   Adversarial Robustness is a growing field that evidences the brittleness of neural networks. Although the literature on adversarial robustness is vast, a dimension is missing in these studies: assessing how severe the mistakes are. We call this notion "Adversarial Severity" since it quantifies the downstream impact of adversarial corruptions by computing the semantic error between the misclassification and the proper label. We propose to study the effects of adversarial noise by measuring the Robustness and Severity into a large-scale dataset: iNaturalist-H. Our contributions are: (i) we introduce novel \textbf{Hierarchical Attacks} that harness the rich structured space of labels to create adversarial examples. (ii) These attacks allow us to benchmark the \textbf{Adversarial Robustness and Severity} of classification models. (iii) We enhance the traditional adversarial training with a simple yet effective \textbf{Hierarchical Curriculum Training} to learn these nodes gradually within the hierarchical tree. We perform extensive experiments showing that hierarchical defenses allow deep models to boost the adversarial Robustness by $1.85\%$ and reduce the severity of all attacks by $0.17$, on average. 
\end{abstract}

\begin{figure*}[t]
    \centering
    \includegraphics[width=0.90\textwidth]{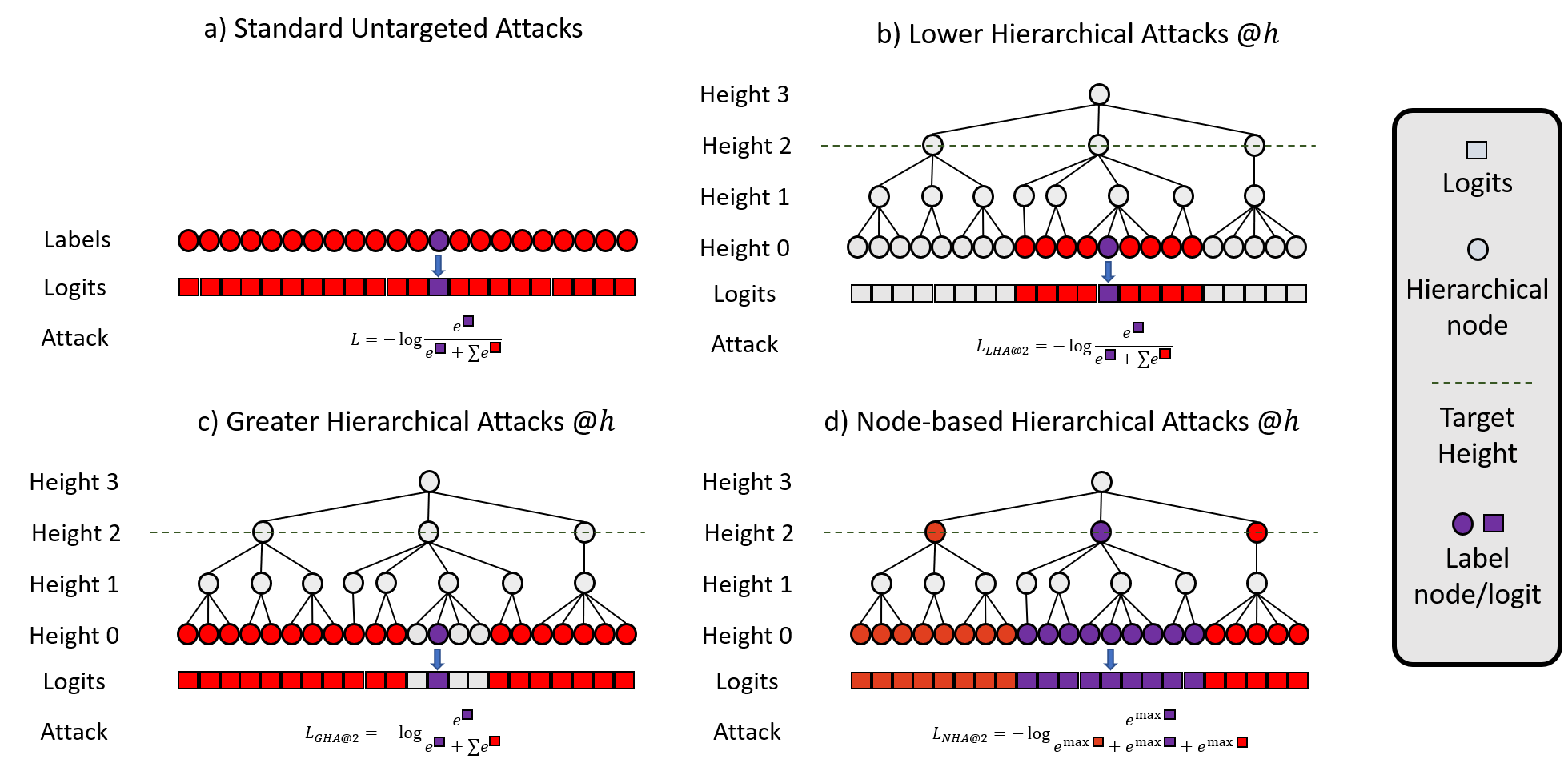}
    \caption{\textbf{Hierarchical Attacks.} In contrast to standard adversarial attacks in (a), we generate adversarial perturbations by considering the hierarchical distance between labels: (b) LHA generates the examples by taking into account only the classes with a hierarchical distance lower or equal than $h$. (c) GHA uses those that are higher or equal than $h$. (d) NHA considers the father nodes with a height level of $l$ as the classification nodes to generate the adversaries. We utilize $h=2$ for the overview of all attacks.} 
    \label{fig:h-attacks}
    \vspace{-3mm}
\end{figure*}

\section{Introduction}

Widely-known adversarial attacks such as FSGM \cite{goodfellow2015explaining}, CW~\cite{carlini2017towards}, Projected Gradient Descent (PGD) \cite{madry2017} or AutoAttack \cite{croce2020reliable}, share a common objective: they aim at decreasing accuracy. While these attacks are effective and practical, they ignore the rich semantic structure of the label space. We hypothesize that disregarding these semantic relations amounts to discarding valuable knowledge that states the degree of relation between classes, and we aim at exploring the notion of \emph{severity} of adversarial attacks. The severity quantifies the semantic error of a misclassification induced by an adversary. The hierarchical distance models this error for a label space formed from a hierarchical tree. The adversarial severity extends the traditional top-1 accuracy over a new dimension. Accordingly, to deploy machine learning algorithms in real-life scenarios, we require them to be protected against adversaries that may exploit the structure of the semantic label space. In particular, models need defenses against attacks capable of radically changing their output. Many cases require to cover a complete range of outputs, where some of them have dissimilar connotations. Take as an example action classification for video surveillance. Changing a prediction ``playing'' to ``running'' is not severe because of its close semantics. Nonetheless, altering ``robbing'' to ``running'' may lead to severe consequences.

This phenomenon is no stranger to DL methods. Since the introduction of DL models, utilizing the hierarchical labels as an additional source of information has been ignored to some extent. Similar to adversarial attacks, these architectures focus exclusively on improving accuracy. Recently, Bertinetto \etal~\cite{bertinetto2020making} studied how modern progress in DL has not translated into improvements in terms of hierarchical error, despite providing significant gains in accuracy. 

In this paper, we quantify the semantic errors in which DL models incur when under attack. This study allows us to identify how current attacks and defenses are insufficient to assess the adversarial severity. Thus, we propose: \textit{(i)} a new set of hierarchy-aware attacks. \textit{(ii)} a new benchmark to assess the adversarial robustness and severity. \textit{(iii)} a defense to diminish the severity and the precision of attacks.

We note that the semantic structure of the label space can guide the design of adversarial attacks. Thus, we study hierarchically-labeled datasets and propose a new set of hierarchy-aware attacks. In contrast to traditional adversarial attacks, the target of these hierarchical attacks is input misclassification and large semantic errors. The suite is composed of three novel attacks: \textit{Lower Hierarchical Attack} (LHA), \textit{Greater Hierarchical Attack} (GHA) and \textit{Node-based Hierarchical attack} (NHA). Figure \ref{fig:h-attacks} illustrates how traditional attacks ignore the rich structure of the label space (Figure~\ref{fig:h-attacks}\textit{a}) and how our attacks exploit this information to induce semantic errors (Figures~\ref{fig:h-attacks}\textit{b-d}). Our attacks employ the hierarchical distance to craft adversaries. Firstly, LHA creates distortions with low hierarchical distances. Secondly, GHA attacks the target image to fool the target network with highly unalike semantics. Finally, NHA crafts adversarial images aiming at changing the parent node.

In order to assess adversarial severity, we develop a comprehensive benchmark built upon the work of \cite{bertinetto2020making}. We use our proposed attacks to diagnose the brittleness of models and the stress under adversarial attacks. The environment given by iNaturalist-H is an appropriate testbed for analyzing both effects, as its rich label space originates from genetics-based phylogenetic trees. So, we measure the Accuracy and the Average Mistake under the influence of our attacks to quantify the adversarial robustness and severity.

To mitigate adversarial severity, we hypothesize that the inclusion of the hierarchy into adversarial training reduces the severity of adversarial attacks. Thus, our approach takes inspiration from curriculum-based human learning \cite{Bengio_Currulum_ICML09, jae11learning}. Intuitively, a student's learning starts from basic and coarse concepts and goes towards finer and specialized ones. This mechanism allows the student to easily learn finer concepts as the coarse notions may be a strong prior. Therefore, we exploit the fact that coarse/fine concepts are naturally defined when an underlying hierarchical structure encompasses the classes. Consequently, we propose a Curriculum for Hierarchical Adversarial Training (CHAT) to gradually learn all nodes at every level in the class hierarchy. Our experiments show that CHAT improves robustness and diminishes the severity against adversaries.

We summarize our contributions as follows:

\begin{itemize}
    \item We introduce a new set of \textit{Hierarchy-aware Attacks} that optimize adversarial examples by aiming at both diminishing the accuracy and increasing hierarchical errors.
    
    \item We provide the \textit{first assessment on adversarial severity} on a highly challenging, large-scale, long-tailed, and hierarchically-structured benchmark: iNaturalist-H.
    
    \item We show how employing \textit{CHAT} to gradually learn classes from a tree results in enhanced defenses against hierarchical adversarial attacks.
\end{itemize}

Our code and models are available at {\tt\small \url{https://github.com/BCV-Uniandes/AdvSeverity}}.

\section{Related Work}

\textbf{Robustness Assessments.} As new attacks are released, new defenses are created and vice-versa, generating an arms race. However, several works have demonstrated that reliably assessing adversarial robustness is an elusive task~\cite{athalye2018obfuscated}. Thus, recent works have tried to provide either formal certifications of robustness~\cite{raghunathan2018certified, cohen2019certified} or reliable benchmarks for empirical assessment of adversarial robustness \cite{dong2020benchmarking, croce2020reliable}. Despite the sophistication of these benchmarks, we observe that they are only concerned with evaluating whether attacks can induce error in network predictions. Thus, these benchmarks disregard how \textit{severe} such errors are. We fill this gap in the literature by introducing novel hierarchy-aware adversarial attacks. Then, we conduct a large-scale evaluation of the severity of errors induced by such attacks in the challenging \textit{iNaturalist-H} \cite{bertinetto2020making, inat19} dataset.

\textbf{Curriculum in Deep Learning.} A plethora of studies show the benefits of using a learning curriculum during training \cite{294207, Bengio_Currulum_ICML09, NIPS2011_f9028fae}. Hacohen and Weinshall \cite{pmlr-v97-hacohen19a} show that progressive learning from easy to hard instances enhances the performance when compared with the standard training. Graves \etal~\cite{pmlr-v70-graves17a} create an automatic curriculum learning to improve convergence time. Weinshall and Amir~\cite{JMLR:v21:18-751} provide a theoretical perspective on curriculum learning. Duan \etal~\cite{10.1007/978-3-030-58598-3_4} explore a curriculum for 3D-shape representation learning. Wug Oh \etal~\cite{stm2019Seoung} use a curriculum on the task of video object segmentation to learn object appearance over long time frames. We propose an adversarial training curriculum that learns concepts from coarse to fine on the hierarchical tree.

\textbf{Hierarchical Classification.} Despite the availability of datasets with hierarchical information~\cite{inat19, NIPS2013_9aa42b31, imagenet_cvpr09, ren2018metalearning}, the computer vision community has not yet standardized the usage of a rich structured space of labels for either training or evaluation. Recently, Bertinetto \textit{et al.}~\cite{bertinetto2020making} conducted the first large-scale evaluation of the Accuracy and the Average Mistakes of modern methods by using the taxonomic trees of iNaturalist \cite{inat19} and tieredImageNet \cite{ren2018metalearning}. 
In~\cite{bertinetto2020making}, Hierarchical-based approaches are split into three groups: Label-embedding methods \cite{8658633, NIPS2013_7cce53cf, pmlr-v37-romera-paredes15, 44873, xian2016latent}, hierarchical losses \cite{bertinetto2020making, bilal2018do, jia2010classifying, verma2012learning} and hierarchical architectures \cite{yolo2017, ahmed2016network, yan2014yan}.
We refer the interested reader to \cite{bertinetto2020making} for a thorough review on hierarchical classification. 
We draw inspiration from hierarchical architectures to introduce a hierarchical curriculum during training to enhance adversarial robustness and reduce the induced semantic error. 

\section{Methods}

\subsection{Notation} \label{sec:notation}

A hierarchical tree with height $H$ is a set of groups of nodes $\{Y_0, Y_1, ..., Y_{H-1}\}$ and their corresponding transition functions $C_{h, h'}$. Each set $Y_h = \{1, 2, ..., n_h\}$ contains all nodes at height $h$, \eg $Y_0$ is the set of leaf nodes and $Y_{H-1}$ is the root. Additionally, the transition function $C_{h, h'}$ maps any label in $Y_h$ into a subset of nodes in $Y_{h'}$, representing the label's offspring when $h>h'$. Formally, $C_{h,h'}: Y_h \rightarrow \mathbb{P}(Y_{h'})$, where $\mathbb{P}(\cdot)$ is the powerset of its input. These transition operations have the following properties:

\begin{enumerate}
    \item $\bigcup_{i \in Y_h} C_{h,h'}(i) = Y_{h'}$
    \item $C_{h,h'}(i) \neq \varnothing$, $\forall i \in Y_h$
    \item $C_{h,h'}(i) \cap C_{h,h'}(j) = \varnothing$, $\forall i, j \in Y_h$, $i \neq j$ if $h \geq h'$.
\end{enumerate}

Let be a dataset with images and their corresponding hierarchical labels from the tree. Let a neural network be a composition between a backbone $g$ that extract features representation of images and a linear classifier $f_{W, b}$, where $W \in \mathbb{R}^{n_0 \times m}$ and $b \in \mathbb{R}^{n_0}$ are the weights and biases of $f$, respectively. $m$ is the feature dimension of the representation vector, and $n_0$ is the number of leaf classes. Define $W_i \in \mathbb{R}^{1 \times m}$ to be the $i$-th row of $W$ and $b_{i} \in \mathbb{R}$ to be the $i$-th component of the bias vector.

To measure distances within the topology of a tree, we use the hierarchical distance $d_H$. Hence, the height of the least common ancestor between nodes defines this metric. So, we use $d_H$ to measure a semantic difference between an instance's prediction and the corresponding ground-truth. Throughout the rest of the manuscript, we refer as \textit{mistake} as the hierarchical distance between two nodes.

\subsection{Hierarchical Attacks}

In order to assess the severity of adversarial attacks, we propose novel hierarchy-aware adversaries that perturb images aiming at reducing standard accuracy and increasing the severity of hierarchical mistakes. Figure \ref{fig:h-attacks} provides a graphical illustration of our new attacks: \textit{Lower Hierarchical Attack}, \textit{Greater Hierarchical Attack} and \textit{Node-based Hierarchical Attack}. As we intend to increase hierarchical mistakes, all our attacks aim to extract and harm the target image by exploiting the rich structure of the label's space hierarchy.
 
\textbf{Lower and Greater Hierarchical Attacks at height $\boldsymbol{h}$.} These attacks aim at perturbing the target image by creating distortions with a criterion based on the hierarchical distance $h$ between the target leaf label and the other leaf nodes. Both attacks operate similarly: to choose the adversary classes depending on the severity of the attack, given by $h$. Please refer to Figure \ref{fig:h-attacks}b and \ref{fig:h-attacks}c for an illustration of the intuition behind our proposed attacks.

On the one hand, the most semantically closed classes to a target label are those whose similarity within the tree is at most $h$ in the hierarchical distance. Accordingly, LHA creates the adversarial examples targeting classes whose distance to the original class is \emph{less} than or equal to $h$. Thus, for an image $x$ and its leaf label $y$, LHA optimizes the loss (Equation \eqref{eq:LHA}):
\begin{equation}\label{eq:LHA}
    \begin{split}
        L_{LHA @ h}(x, y)& = \\
        -\log &\frac{e^{W_y g(x) + b_y}}{\sum_{j \in Y_0 \;\text{s.t.}\; d_H (y,j) \leq h} e^{W_j g(x) + b_j}}.
    \end{split}
\end{equation}

On the other hand, GHA@$h$ aims at creating adversaries that induce large mistakes (\textit{i.e.} cause remarkable ``confuse'' of the network). Conversely to LHA, GHA uses information from the classes with a hierarchical distance \textit{greater} or equal to $h$. Thus, GHA maximizes the loss (Equation \eqref{eq:GHA}):
\begin{equation}\label{eq:GHA}
    \begin{split}
        L_{GHA @ h}(x&, y) = \\
         -\log & \frac{e^{W_y g(x) + b_y}}{\sum_{j \in \{k \in Y_0 | d_H(y,k) \geq h \text{ or } k=y \} } e^{W_j g(x) + b_j}}.
    \end{split}
\end{equation}

To ease the understanding of these losses, we point out that the objective to minimize is similar for both attacks. On Figure \ref{fig:h-attacks}b and \ref{fig:h-attacks}c, in essence, both attacks harms the purple node -the label- by using as negative classes the red leaves.

\textbf{Node-based Hierarchical Attacks.} The NHA harms the nodes directly with a height of $h$. Figure \ref{fig:h-attacks}d exemplifies our proposed attack. To compute the probability of classifying a target image $x$ as a node $y$ with height $h$, we follow the conditional probability theory. Thus, this probability equals the sum of the probabilities of choosing any leaf class derived from $y$. Therefore,
\begin{equation}\label{eq:prob-of-node}
    P(c(x)=y \in Y_h) = \smashoperator{\sum_{i \in C_{h,0}(y)}} P(c(x) = i \in Y_L)
\end{equation}
where $c(x) = y$ is the action of classifying $x$ as $y$ and $P(c(x)=y \in Y_0)$ is the probability of classifying $x$ as $y \in Y_0$. As is standard, all state-of-the-art models use a Softmax layer to compute the probability distribution of an input image by using the corresponding logits. Therefore, the probability of the neural network to classify $x$ as class $i$ is:
\begin{equation}
    P(c(x)=i \in Y_0| W, b) = \frac{e^{W_i g(x) + b_i}}{\sum_{j=1}^{n_0}e^{W_j g(x) + b_j}}.
\end{equation}
Hence, following Equation \eqref{eq:prob-of-node}, the probability of classifying $x$ as $y \in Y_h$ is:
\begin{equation} \label{eq:prob-of-node-model}
    \begin{split}
        P(c(x)=y \in Y_h | &W, b) =\frac{\sum_{i \in C_{h,0}(y)}e^{W_i g(x) + b_i}}{\sum_{j=1}^{n_0}e^{W_j g(x) + b_j}}.
    \end{split}
\end{equation}

To extract adversaries based on Equation \eqref{eq:prob-of-node-model}, we compute the cross-entropy loss. To speed up this operation, we estimate the adversaries by attacking the cross-entropy on Equation \eqref{eq:NHA}. This function computes the loss over the maximum logits within nodes of interest:
\begin{equation} \label{eq:NHA}
    \Tilde{L}_{NHA@h}(x, y) = -\log \frac{e^{\Tilde{L}_{y}}}{\sum_{j=1}^{n_h} e^{\Tilde{L}_j}}
\end{equation}
where, $\Tilde{L}_y = \max_{i \in C_{h,0}(y)} W_i g(x) + b_i$. We base our design on the mathematical principle that, if $L_y$ is the logit corresponding to node $y \in Y_h$, with straightforward algebraic manipulation we can obtain that
\begin{equation} \label{eq:logit-node}
    L_y = \log \left[ \smashoperator[r]{\sum_{i \in C_{h,0}(y)}}e^{W_i g(x) + b_i}\right].
\end{equation}

It is widely known that the $\max$ function has a variety of differentiable approximations. We recall one in particular: $max (x_1, ..., x_m)
\approx \log (e^{x_1} + ... + e^{x_m})$. Thus, we find that $L_y \approx \Tilde{L}_y$. As a result, we attack an approximation of the true cross-entropy function--Equation \eqref{eq:NHA}. In Figure \ref{fig:h-attacks}d, we exemplify the idea behind NHA. This attack extracts the maximum logit over the leaf offspring of each node at the target height. So, NHA harness these logits to create the adversary.

\subsection{Benchmarking Adversarial Severity}

In order to conduct an assessment of both adversarial robustness and severity, we compute two metrics that reflect these concepts. First, we compute the standard \textit{Robust Accuracy}. This measurement quantifies the worst-case endurance of a defense against adversarial examples. However, this metric does not provide information about the semantic error the model incurred under the attack. Hence, we also compute the \textit{Average Mistake} \cite{bertinetto2020making}. So, we average the hierarchical distance $d_H$ of all misclassified instances. This metric quantifies the semantic dissimilarity between the prediction and the ground-truth label, thus, addressing the issue presented by only measuring accuracy.

Fundamentally, our attacks propose novel optimization objectives. In practice, for optimizing such objectives, we implement a PGD-based strategy. The standard PGD~\cite{madry2017} iteratively performs FSGM steps \cite{goodfellow2015explaining} to maximize the Cross-Entropy Loss $L$ between the prediction for an instance $x$ and its corresponding ground-truth label $y$ to find an adversarial example:
\begin{equation}\label{eq:pgd-attack}
    \begin{split}
    x_{t+1} = \smashoperator[r]{\prod_{B_{\ell_\infty}(x, \epsilon)}} x_t + \alpha\:\text{sign}\left(\nabla_{x_t} (L(m(x_t), y))\right),
    \end{split}
\end{equation}
where $m(\cdot) = Softmax(f_{W, b}(g(\cdot)))$, and the initial point $x_0$ is perturbed with noise, namely $x_0 = x + u$, with $u$ being sampled from the uniform random distribution $\mathcal{U}[-\epsilon, \epsilon]$, $\alpha$ is the step per iteration and $\prod_{B_{\ell_\infty}(x, \epsilon)}$ is the projection function over the set $B_{\ell_\infty}(x, \epsilon)$, namely, the intersection set between the $\epsilon$-ball with the $\ell_\infty$ norm around $x$ and the set $[0,1]^{size(x)}$. 
We refer to an $n$-step PGD attack as PGD\textbf{$n$}. 
Using PGD-optimization for our attacks amounts to replacing the Loss $L$ in Equation \eqref{eq:pgd-attack} with each of the losses in Equations \eqref{eq:LHA}, \eqref{eq:GHA} and \eqref{eq:NHA}. 
Analogously, we refer to the $n$-step versions of these attacks as LHA\textbf{$n$}@$h$, GHA\textbf{$n$}@$h$ and NHA\textbf{$n$}@$h$. For all our evaluation experiments, we set $\alpha=\nicefrac{1}{255}$ and report the worst-case accuracy. 
Furthermore, we compute the adversaries for correctly-classified instances.

\textbf{Dataset:} We construct our methodology on top of the \textit{iNaturalist-H}, created by \cite{bertinetto2020making}. This dataset is a partition of the challenging iNaturalist 19 \cite{inat19}. It is known by how its class distribution follows a long-tail, resembling the imbalance of the real world. This dataset covers a total of 1010 leaf classes dominated by an 8-level phylogenetic tree. We perform all our ablation experiments on the validation set and report the main results in Figure \ref{fig:main_results} on the test set. We provide a detailed description of this dataset and useful statistics on the \textbf{Supplemental Material}.

\subsection{Curriculum for Hierarchical Adversarial Training}

\begin{figure}[t]
    \centering
    \includegraphics[width=0.90\columnwidth]{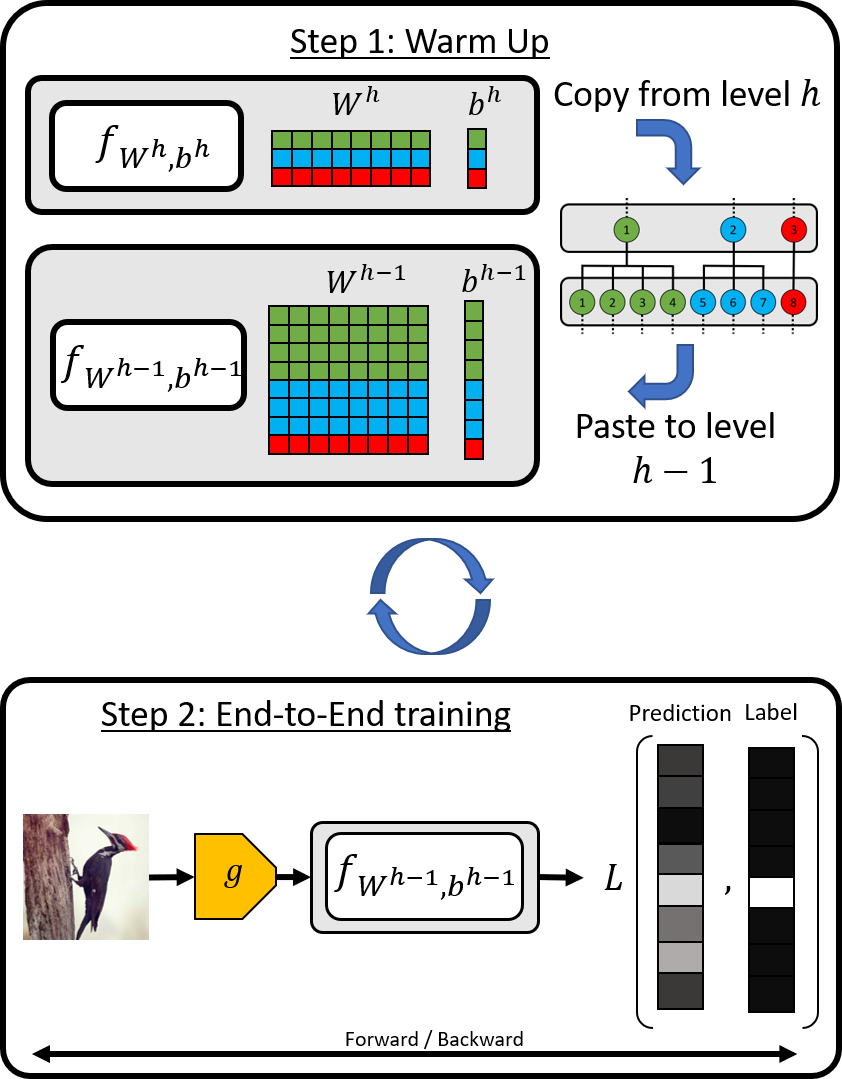}
    \caption{\textbf{Overview of CHAT.} Our hierarchical curriculum has two steps: the warm up (upper Figure) and the end-to-end training (bottom Figure). The warm up consist on transferring the weights and biases from the parent node to its offspring. The second step consist on training the model's parameters, $W^{h-1}$ and $b^{h-1}$ in a end-to-end manner.
    }
    \label{fig:curriculum}
    \vspace{-3mm}
\end{figure}

In order to defend our model against accurate and severe attacks, we propose CHAT: Curriculum for Hierarchical Adversarial Training. Inspired by human learning, we create a curriculum-based training to learn all the nodes by progressively increasing the difficulty of the task, \textit{i.e.} iteratively deepening the current height $h$ of tree to $h-1$ at each stage, until reaching $h=0$. Our curriculum comprises two iterative steps: the warm up and the end-to-end training. Figure \ref{fig:curriculum} illustrates a stage on our curriculum training. For the first step, consider $W^h$ and $b^h$ to be the weights and biases of the current classifier at the $h$-th stage of the curriculum. When updating the step from $h$ to $h-1$, the size of the weights of $f$ must change to account for the increase in the number of classes. Thus, we initialize a new classifier $f_{W^{h-1}, b^{h-1}}$, with $W^{h-1} \in \mathbb{R}^{n_{h-1} \times m}$ and $b^{h-1} \in \mathbb{R}^{n_{h-1}}$. Then, we provide a warm up for the weights $W^{h-1}$ and bias $b^{h-1}$ by transferring the parameters from each node $j$ to all its children. Finally, we discard the weights $W^h$ and biases $b^h$. Formally, for all $i \in C_{h,h-1}(j)$ we apply:
\begin{equation}
    \begin{split}
        W_i^{h-1} &= W_j^h \\
        b_i^{h-1} &= b_j^h
    \end{split}
\end{equation}
Due to properties 1, 2 and 3 of the transition operations in section \ref{sec:notation}, we ensure \textit{(i)} that there is always at least an offspring at each step, \textit{(ii)} that there are no child nodes that descend from two parents, and \textit{(iii)} we cover all nodes at height $h-1$. The second step consists of training all parameters of both the feature extractor and the linear classifier, $W^{h-1}$ and $b^{h-1}$, in an end-to-end manner. The optimization process uses the labels from the current stage, namely $Y_{h-1}$, to minimize the cross-entropy loss. Furthermore, we robustify the model on this step by replacing the training with any adversarial learning method \cite{Wang_2019_ICCV, free2019, madry2017}. We iterate these stages until convergence at the leaf nodes. We enforce the number of epochs at each stage is closely related to the number of nodes at each level. 

To train our models to be robust, we adopt an Adversarial Training~\cite{madry2017} inspired strategy. In particular, we follow the computationally-efficient \textit{Free Adversarial Training} (FAT) technique \cite{free2019}, which exploits each forward-backward pass to optimize both model parameters and the adversarial examples used for training. FAT has three hyper-parameters of interest: the number of times each adversarial example is replayed, $m$, the bound for the adversarial examples, $\epsilon$, and the adversary optimization's steps size, $\alpha$.

 \section{Experiments}

 \begin{figure*}[t]
     \centering
     \begin{subfigure}[b]{0.33\textwidth}
        \includegraphics[width=\textwidth]{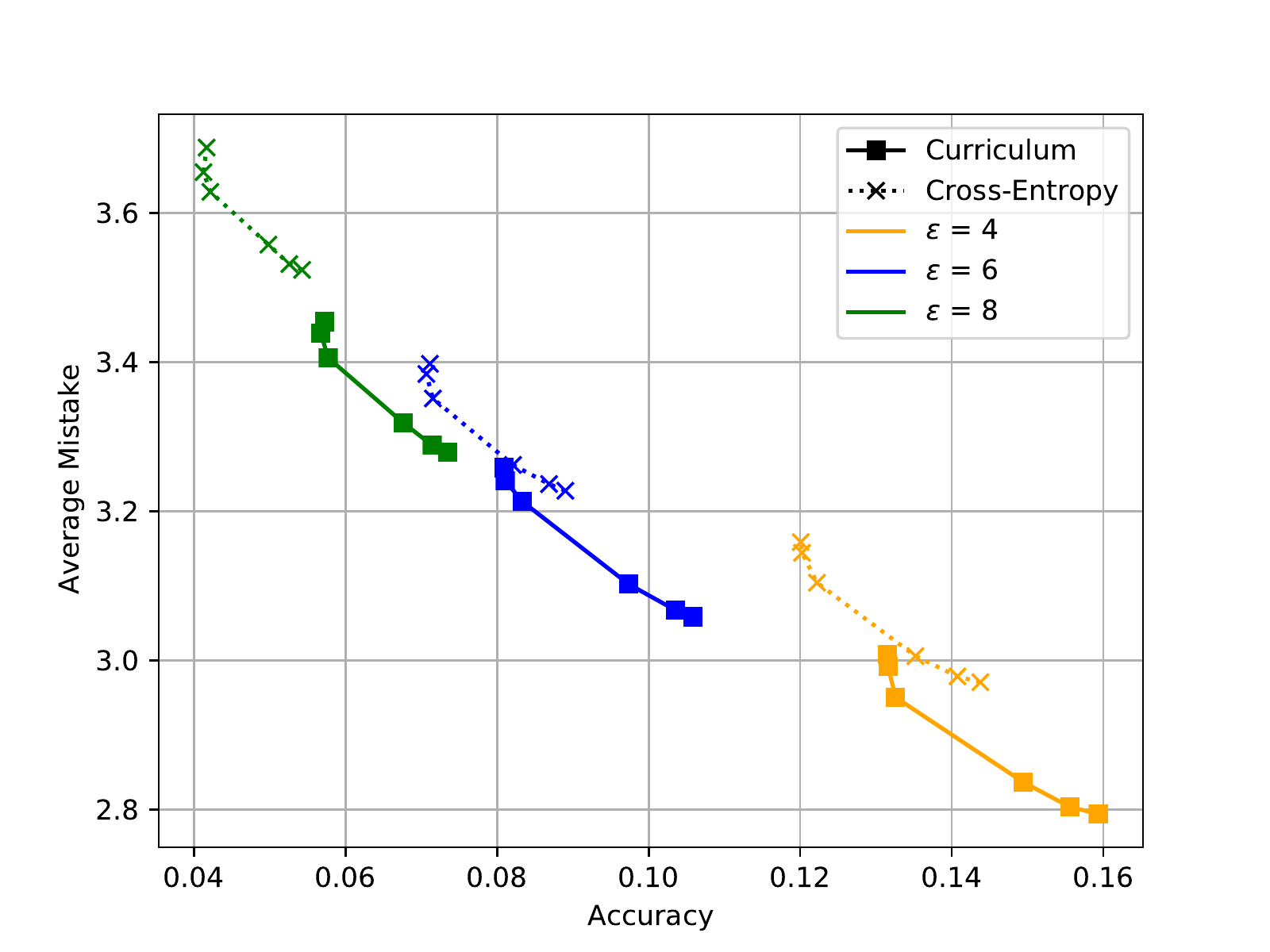}
        \caption{LHA Assessment}
     \end{subfigure}
     \hfill
     \begin{subfigure}[b]{0.33\textwidth}
        \includegraphics[width=\textwidth]{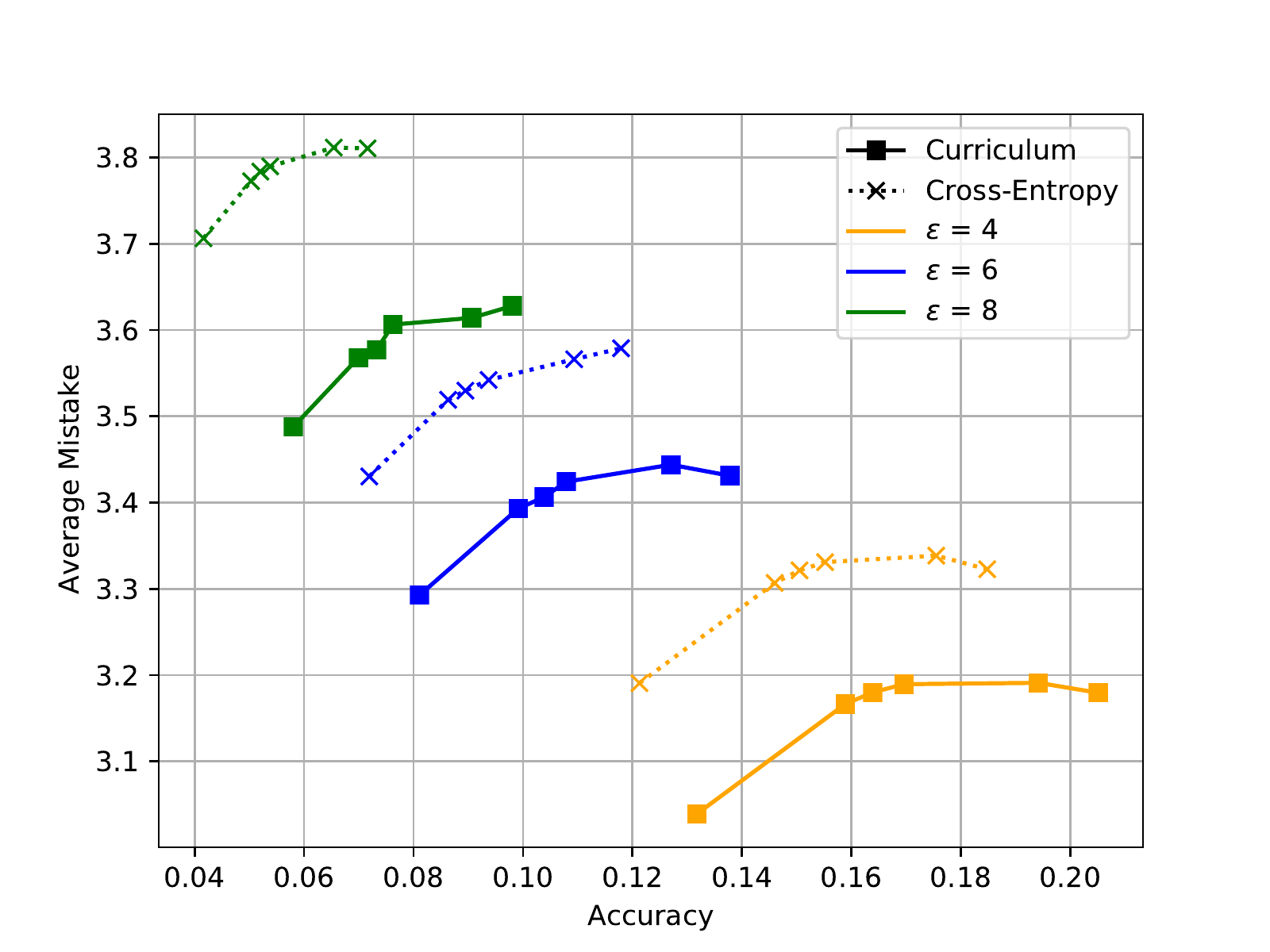}
        \caption{GHA Assessment}
     \end{subfigure}
     \hfill
     \begin{subfigure}[b]{0.33\textwidth}
        \includegraphics[width=\textwidth]{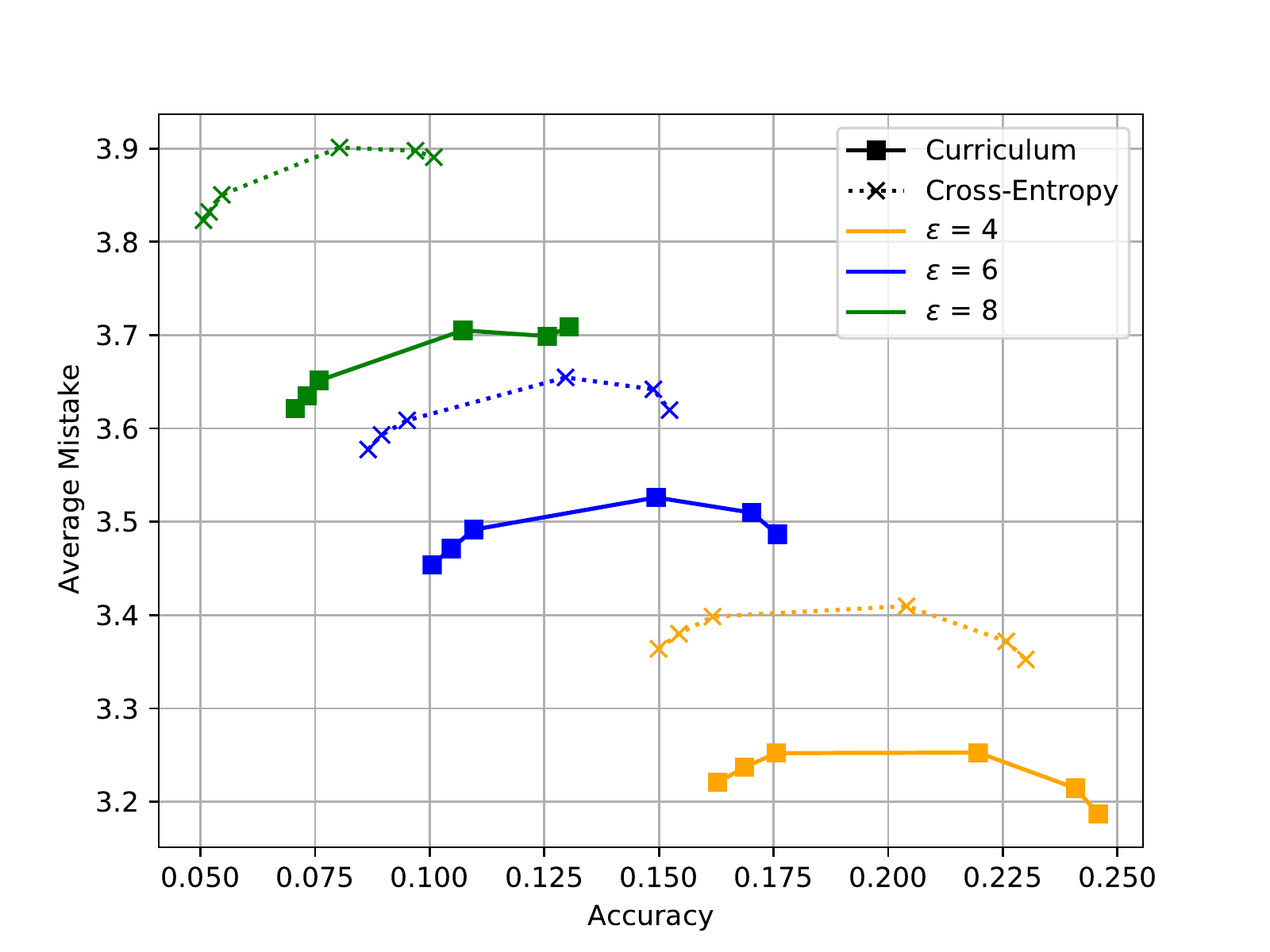}
        \caption{NHA Assessment}
     \end{subfigure}
     \caption{\textbf{Hierarchical Attack Evaluation.} We evaluate the performance of FAT on our hierarchical attacks with 50 iterations at each target height on the test set. The inclusion of our curriculum boost all metrics for all levels compared to the baseline. (a) LHA is the strongest accuracy-related attack among the proposed ones. It even surpasses the PGD when setting the height $h$ to 5 or 6. (b) The GHA enjoys a balanced between severity and accuracy. (c) The NHA is the most severe but less successful attack among the set of hierarchical attacks.}
     \label{fig:main_results}
     \vspace{-0.3cm}
 \end{figure*}

\begin{table}[t]
    \centering
    \begin{tabular}{c|cc|c|cc|cc} \toprule
    \multirow{2}{*}{$\boldsymbol{\epsilon}$} & \multirow{2}{*}{$\boldsymbol{m}$} & \multirow{2}{*}{$\boldsymbol{\alpha}$} & \multirow{2}{*}{\textbf{C}} & \multicolumn{2}{c|}{\textbf{Clean}} & \multicolumn{2}{c}{\textbf{PGD50}}\\
                    &&&& \textbf{Acc}   & \textbf{AM}    & \textbf{Acc}   & \textbf{AM}    \\ \midrule
        4 & 6 & 6 &            & 31.40 & 3.21  & 12.33 & 3.20  \\
        4 & 8 & 6 & \checkmark & 32.84 & 3.04  & 13.36 & 3.06  \\ \midrule
        6 & 6 & 4 &            & 24.87 & 3.44  & 7.13  & 3.44  \\
        6 & 8 & 4 & \checkmark & 27.19 & 3.28  & 8.32  & 3.29  \\ \midrule 
        8 & 6 & 6 &            & 19.65 & 3.76  & 4.31  & 3.71  \\
        8 & 8 & 6 & \checkmark & 23.29 & 3.49  & 6.07  & 3.49 \\\bottomrule 
    \end{tabular}
    \vspace{2mm}
    \caption{\textbf{Clean and PGD50 performaces for the best models.} We compare the best models we extracted through the grid search on the space of hyperparameters. C stands for curriculum, $m$ are the number of iterations per images and $\alpha$ the step size. The results show that our proposed curriculum greatly enhances the performance on all metrics by using FAT.}
    \label{tab:best-models}
    \vspace{-3mm}
\end{table}

\subsection{Curriculum-Enhanced Models against Hierarchical Attacks}

\textbf{Implementation Details.} For all our experiments we follow the experimental setup of \cite{bertinetto2020making}. We initialize a ResNet-18 \cite{resnet16} from ImageNet-pretrained \cite{imagenet_cvpr09} weights and then train the network for $200,000$ steps. We use the Adam optimizer \cite{adam2018} with $\beta_1=0.9$ and $\beta_2=0.999$, learning rate of $10^{-5}$, no weight decay, and a mini batch-size of $256$. We fix the updates in curriculum stages to occur at $2\%$, $4\%$, $6\%$, $15\%$, $25\%$ and $35\%$ of the training iterations.

\textbf{Model Selection.} We assess the effects of our new attacks against our CHAT-enhanced models. We selected the models that provided the best robustness against PGD50 on the validation set for a fair comparison between the vanilla and CHAT-enhanced models. We explore the space of parameters $m \in \{ 2, 4, 6, 8\}$ and $\alpha\in \{\nicefrac{1}{255}, \nicefrac{2}{255}, \nicefrac{4}{255}, \nicefrac{6}{255}\}$ for the standard and hierarchy-enhanced models to search for the best-performing models for all $\epsilon \in \{\nicefrac{4}{255}, \nicefrac{6}{255}, \nicefrac{8}{255}\}$. We report each model's hyperparameters and performance on the validation set in Table \ref{tab:best-models}. The results show that the proposed CHAT enhances \textit{all} metrics on the clean and PGD50 settings by a large margin. We underscore that all curriculum-enhanced models were trained with a higher number of iterations $m$. We suspect that a higher number of iterations per height will increase even further the gap. Nonetheless, a complete study on the curriculum aspect of the training is out of the scope of this paper. Also, the step sizes of the models are large in comparison to their perturbation budget. Wang and Zhang~\cite{Wang_2019_ICCV} evidence that some models enjoy similar robustness for larger step sizes and fewer iterations when trained adversarially on the CIFAR dataset. Although we are tackling a higher-dimensional dataset, we observe this phenomenon too. FAT approximates a 1-step PGD training routine. Thus, large step sizes may further enhance adversarial robustness.

\begin{table}[t]
    \centering
    \begin{tabular}{c|cccc}\toprule
        \textbf{Metric}   & \textbf{PGD5}  & \textbf{PGD10} & \textbf{PGD50} & \textbf{PGD100} \\ \midrule
        Accuracy & 13.85 & 11.44 & 11.07 & 11.06  \\
        AM       & 3.25  & 3.25  & 3.25  & 3.25   \\ \bottomrule
    \end{tabular}
    \vspace{2mm}
    \caption{\textbf{PGD iterations.} We assess the number of iterations under the same PGD attack for a vanilla ResNet-18 trained with Cross-Entropy. This experiment shows that PGD50 explores close results as the PGD100.}
    \label{tab:pgd-iterations}
    \vspace{-3mm}
\end{table}

Once we choose all the best models, we quantify the effectiveness of our attacks on the testing set. Figure~\ref{fig:main_results} report the results for LHA50@$h$, GHA50@$h$ and NHA50@$h$ in the test set. These results demonstrate that introducing CHAT improves both the adversarial accuracy and severity metrics for \textit{all} models. The results of LHA present an average increase of $1.45\%$ for the robust accuracy and an average mistake decrease of $0.18$. The GHA and NHA present an average boost on the robust accuracy of $1.72\%$ and $1.91\%$, respectively. Furthermore, The average mistake metric decreases by $0.16$ for GHA and NHA. We attribute these gains to the curriculum's semantic partition, which demonstrates that semantically-coherent representation spaces enhance model robustness.

We further observe that the average mistake saturates at heights greater than or equal to 4. In contrast, the accuracy does not reach such a plateau. This event seems contradictory to the result of Figure \ref{fig:ha-ablations}. Nonetheless, the average mistake metric on this table considers both instances that initially were erroneous and not perturbed and those whose attack was successful.

\subsection{Ablation Studies}

This section studies \emph{(i)} the proposed defense mechanism's components and \emph{(ii)} the effects of the hierarchical-aware attacks. We perform all ablation experiments in the validation set. From the side of the attacks, we first explore the effect of the number of iterations for PGD and report our results in Table \ref{tab:pgd-iterations}. From these results, we conclude that having $50$ iterations is reasonable for a clear evaluation. Thus, for all evaluations of adversarial robustness, we report numbers with respect to PGD50. We study two factors of interest from the training side: the curriculum and the number of iterations between curriculum stages. We train the ResNet-18 with a perturbation budget of $\epsilon=4$, a step size of $\alpha = 2$ and $m=8$ iterations to evaluate all our ablations.

\begin{table}[t]
    \centering
    \begin{tabular}{c|cc|cc}\toprule
        \multirow{2}{4em}{\textbf{Method}} & \multicolumn{2}{c|}{\textbf{Clean}} & \multicolumn{2}{c}{\textbf{PGD50}} \\
                         & \textbf{Acc} & \textbf{AM} & \textbf{Acc} & \textbf{AM}\\ \midrule
        Standard   & 29.64          & 3.25          & 11.07          & 3.25\\
        Scratch         & \textit{34.02} & \textit{3.04} & \textit{11.90} & \textit{3.05}\\
        CHAT (Ours)            & \textbf{34.11} & \textbf{2.98} & \textbf{12.15} & \textbf{3.01}\\ \bottomrule
    \end{tabular}
    \vspace{2mm}
    \caption{\textbf{Effects of CHAT on Adversarial Training.} We report the results of Clean and PGD50 attack accuracy (Acc) and Average Mistake (AM). 
    Results in \textbf{bold} and \textit{italic} show the best and second best performances, respectively.}
    \label{tab:curr_ablation}
    \vspace{-3mm}
\end{table}

\textbf{Weight Transfer.} At each stage of the curriculum, the size of the linear classifier $f$ increases, implying that new weights are required. Here we assess the impact of employing the warm-up stage in contrast to \textit{(i)} employing no curriculum (Standard) and \textit{(ii)} a naive initialization from scratch at each curriculum stage (Scratch). We report the results of this experiment in Table \ref{tab:curr_ablation}. Our results show that the sole inclusion of hierarchical information into FAT (``Scratch'' \textit{vs.} ``Standard'' in Table \ref{tab:curr_ablation}) greatly improves accuracy and diminishes the error severity on both clean and PGD50 settings. Moreover, employing our warm-up strategy when changing between hierarchical levels boosts performance metrics (``CHAT (Ours)'' \textit{vs.} ``Scratch'' in Table \ref{tab:curr_ablation}). Since the Cross-Entropy ignores the relationship between labels, semantically dissimilar classes may be adjacent in the representation space. Although previous works \cite{bilal2018do} show that Cross-Entropy is capable of learning class hierarchies, our experiments suggest that employing a curriculum encourages semantically-similar classes to lie closer in representation space. We argue that this may be because they have a disjoint hyperspace for coarse classes as a prior, learned previously in the curriculum stages.

\begin{table}[t]
    \centering
    \begin{tabular}{c|cc|cc} \toprule
        \multirow{2}{*}{\textbf{Curriculum}} & \multicolumn{2}{c|}{\textbf{Clean}} & \multicolumn{2}{c}{\textbf{PGD50}} \\
                       & \textbf{Acc}  & \textbf{AM}   & \textbf{Acc}   & \textbf{AM}\\ \midrule
        Linear        & 26.39          & 2.97          & 10.52          & 3.04\\
        Ours          & 34.11          & 2.98          & 12.15          & 3.01\\ \midrule
        Change        & \textbf{+7.72} & \textit{+0.01}& \textbf{+1.63} & \textbf{-0.03} \\\bottomrule
    \end{tabular}
    \vspace{2mm}
    \caption{\textbf{Effect of the Curriculum Steps.} We test two different curriculums and report both accuracy (Acc) and Average Mistake (AM) for clean and PGD50 adversaries. We test an exponential-like curriculum, labelled as ours, and a linear one. Given the exponential growth of the number of nodes at each level of the hierarchy tree, an exponential-like curriculum fits perfectly. \textbf{Bold} changes show gains and in \textit{italic} losses.}
    \label{tab:curr_spacing}
    \vspace{-3mm}
\end{table}

\textbf{Step Spacing.} The number of iterations used during parameter optimization can have large consequences on final performance. Since a curriculum on a hierarchy implies a changing number of classes at each stage, the number of optimization iterations run at each stage becomes an influential factor in performance. Thus, we assess the effect of the amount of iterations to optimize each curriculum stage. 
In particular, we test two spacing strategies between curriculum stages: a naive linear spacing and our proposal, an exponential-like spacing. We report our results on Table~\ref{tab:curr_spacing}. These results suggest that providing an exponentially-increasing number of iterations for optimizing the classes at each height of the tree leads to higher-performing \textit{and} more robust models. Combining these results with the evidence from Table \ref{tab:curr_ablation}, we argue that inducing hierarchy-aware priors is key to training more accurate and robust models. Using a linear spacing, we enforce this phenomenon: the small difference on Average Mistake values shows a semantically disjoint space for the parent nodes. Nonetheless, the amount iterations available on the last stage does not enable the convergence into a local minimum for the leaf nodes.

On the \textbf{Supplemental Material} we present further experimentation with TRADES~\cite{zhang2019theoretically} and TRADES enhanced with CHAT.

\begin{figure}[th]
    \centering
    \includegraphics[width=\columnwidth]{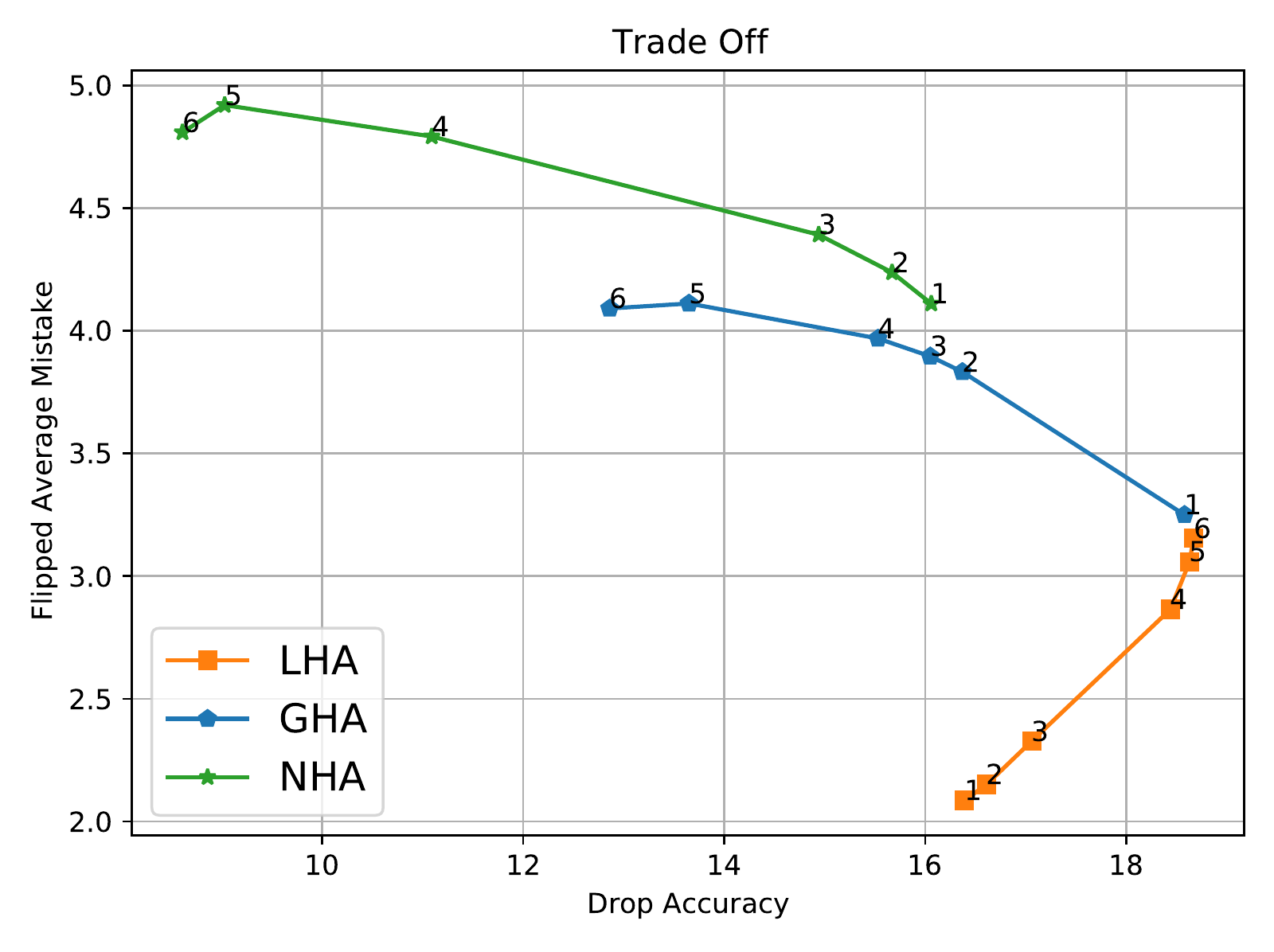}
    \caption{\textbf{Hierarchical Attack effects.} We plot the Difference between Clean and Robust Accuracy (``Accuracy Drop'') against the flipped Average Mistake. The LHA generates semantically similar variances to each instance, while GHA creates dissimilar noise. The NHA creates weak but severe attacks as the curve is below the other ones. The number on top of each point of the curve represents the height of the hierarchical attack.}
    \label{fig:ha-ablations}
    \vspace{-3mm}
\end{figure}

\textbf{Hierarchical Attacks.} Recall that our proposed hierarchy-aware attacks have different objectives in mind, all related to inducing mistakes of varying severities. In order to visualize each effect, we plot on Figure~\ref{fig:ha-ablations} the trade-off between \textit{(i)} the Accuracy Drop (``Clean Accuracy'' minus ``Robust Accuracy'') against \textit{(ii)} the average mistake of those instances whose the attack was successful (``Flipped Average Mistake''). The former dimension looks at how powerful an attack is, and the latter reviews the severity of the successful attack. The results show the intended effects of each attack: The LHA and GHA produce inverse effects, the former being semantically similar and stronger attacks, and the latter more severe but less effective attacks. Note that GHA@1 is equal to standard PGD. We expected the standard PGD to outperform all other attacks in the accuracy metric because it has a broader range of information. To our surprise, both LHA@6 and LHA@5 achieve slightly larger accuracy drops of $0.09$ and $0.05$ points, respectively. We attribute this result to the existence of useless gradients that hinder the effectiveness of gradient-based attacks, as observed by previous studies \cite{pmlr-v80-athalye18a, papernot2017}. Finally, the most severe attack, in terms of semantic mistakes, is our newly proposed NHA. On average, we find average mistakes of at least 0.86 and up to 1.67 points compared to the standard PGD50. Note that the proposed attacks are not as effective at degrading accuracy compared to the standard PGD. Since some images may contain semantic regions similar to some classes, the PGD attack covers this spectrum of ranges. In contrast, our proposed attacks have a reduced sight of these classes. On the \textbf{Supplemental Material} we present further experimentation mixing AutoAttack~\cite{croce2020reliable} with NHA.

\subsection{Qualitative Results}

\begin{figure}
    \centering
    \begin{subfigure}[b]{0.90\columnwidth}
       \includegraphics[width=\textwidth]{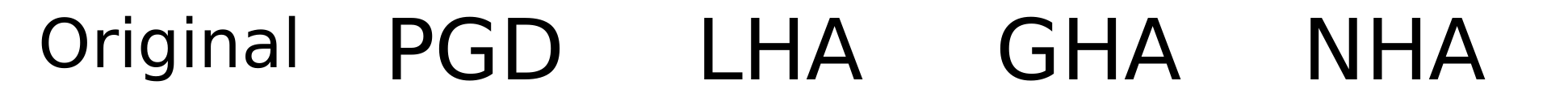}
    \end{subfigure}
    \begin{subfigure}[b]{0.90\columnwidth}
       \includegraphics[width=\textwidth]{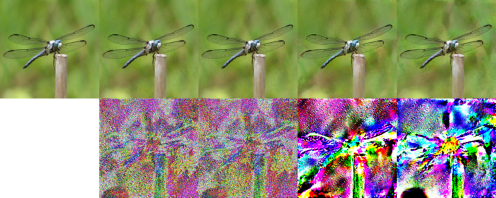}
    \end{subfigure}
    \begin{subfigure}[b]{0.90\columnwidth}
       \includegraphics[width=\textwidth]{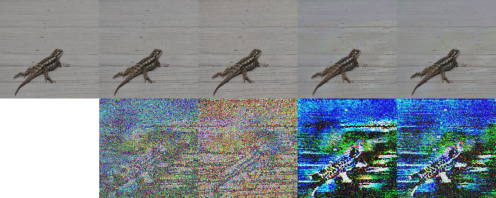}
    \end{subfigure}
    \begin{subfigure}[b]{0.90\columnwidth}
       \includegraphics[width=\textwidth]{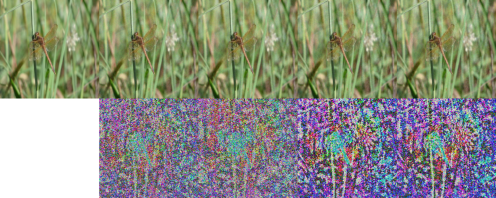}
    \end{subfigure}
    \caption{\textbf{Adversarial Examples.} We visualize some adversarial examples with $\epsilon = \nicefrac{8}{255}$. From left to right, we display the original image, PGD, LHA@$3$, GHA@$3$ and NHA@$3$. Each image bellow the adversarial example is the corresponding adversarial noise. All proposed attacks explore the semantics within the image in different manners.}
    \label{fig:adv_examples}
    \vspace{-3mm}
\end{figure}

We visualize the effects of the proposed attacks and PGD adversaries on Figure \ref{fig:adv_examples}. We used our CHAT model to compute the adversaries. We set the height $h$ to 3, and $\epsilon = \nicefrac{8}{255}$. In addition, we visualize their corresponding perturbation for all adversarial examples. In the first set of attacks, we notice that PGD and LHA behave similarly, while the GHA and NHA noises followed different directions, as noted by the color of the noise. These results exemplify that the PGD attack and the LHA use the same information to create the adversaries for this instance, while the NHA and GHA used different gradients to reach their local minimum. Contrastively, the PGD and LHA noises behave differently in the second set of adversarial examples; the color of both noises differs. Nonetheless, the GHA and NHA create similar perturbations. Finally, the last instance shows that all attacks may similarly create adversaries. We set further visualizations of multiple adversarial examples under different depths on the \textbf{Supplemental Material}.

\section{Conclusions}

In this paper, we unravel the rich semantic structure of the label space to devise a new set of hierarchical attacks. To assess their effects, we extend the classical evaluation metric of the adversarial accuracy and explore a new dimension of adversarial attacks: their severity. Consequently, we exploit iNaturalist-H, a large-scale dataset with a label space generated from a taxonomic tree, and create a benchmark with the aforementioned metrics. Furthermore, we propose CHAT, a curriculum-enhanced training to improve the robustness against adversarial examples and the severity of the damage by using all the hierarchical nodes of the taxonomic tree. We hope that studying adversarial severity opens new research directions in robustness.

{\small
\bibliographystyle{ieee_fullname}
\bibliography{egbib}
}

\newpage

\textbf{{\Large Supplemental Material}}

\section*{iNaturalist-H Statistics}

We present the statistics of iNaturalist-H on table \ref{tab:stats}. This dataset contains 189404 images for training, 42140 for validation, and 42756 for testing. The dataset includes a broad span of classes, ranging from animals to plants. The imbalance is extremely high: the least number of images per category is 13, and the maximum is 352. Furthermore, the standard deviation of the number of leaf nodes per father node is enormous, showing imbalance even on supernodes.

\begin{table}[h]
    \centering
    \begin{tabular}{c|ccc} \toprule
        \textbf{Level}   & \textbf{Mean}   & \textbf{std}    & \textbf{Nodes}\\ \midrule
        Kingdom & 336.67 & 273.92 & 3\\
        Phylum  & 252.50 & 255.32 & 4\\
        Class   & 112.22 & 169.45 & 9\\
        Order   & 29.71  & 17.49  & 34\\
        Family  & 17.72  & 9.73   & 57\\
        Genus   & 14.02  & 4.95   & 72\\
        Species & 1      & 0      & 1010\\ \bottomrule
    \end{tabular}
    \vspace{2mm}
    \caption{\textbf{Node statistics of iNaturalist19.} The iNaturalist presents a high imbalance on both number of instances per class and number of leaf classes descendant from the nodes. Also we present the number of nodes on each level.}
    \label{tab:stats}
\end{table}

 \begin{figure*}[t]
     \centering
     \begin{subfigure}[b]{0.33\textwidth}
        \includegraphics[width=\textwidth]{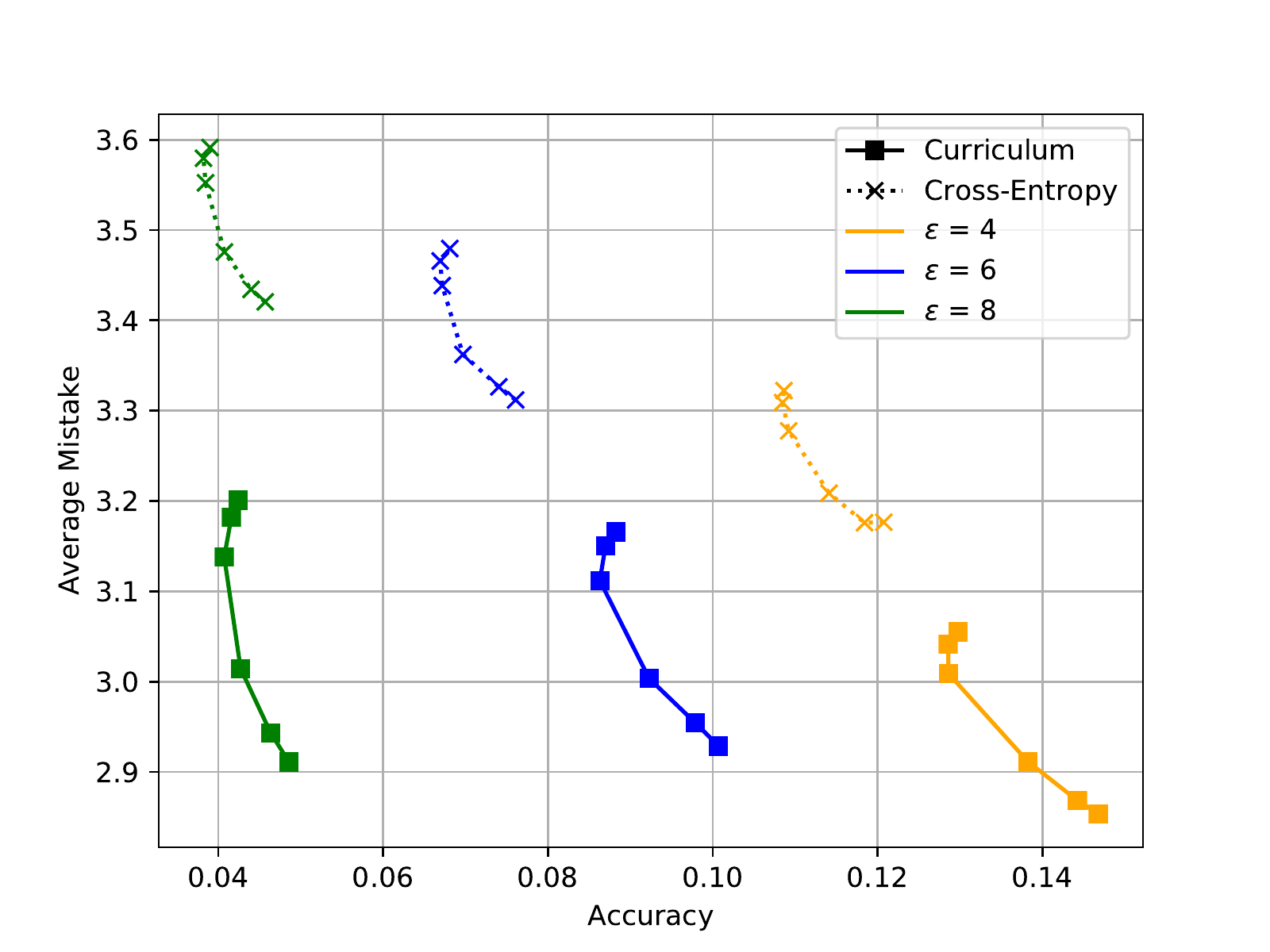}
        \caption{LHA Assessment}
     \end{subfigure}
     \hfill
     \begin{subfigure}[b]{0.33\textwidth}
        \includegraphics[width=\textwidth]{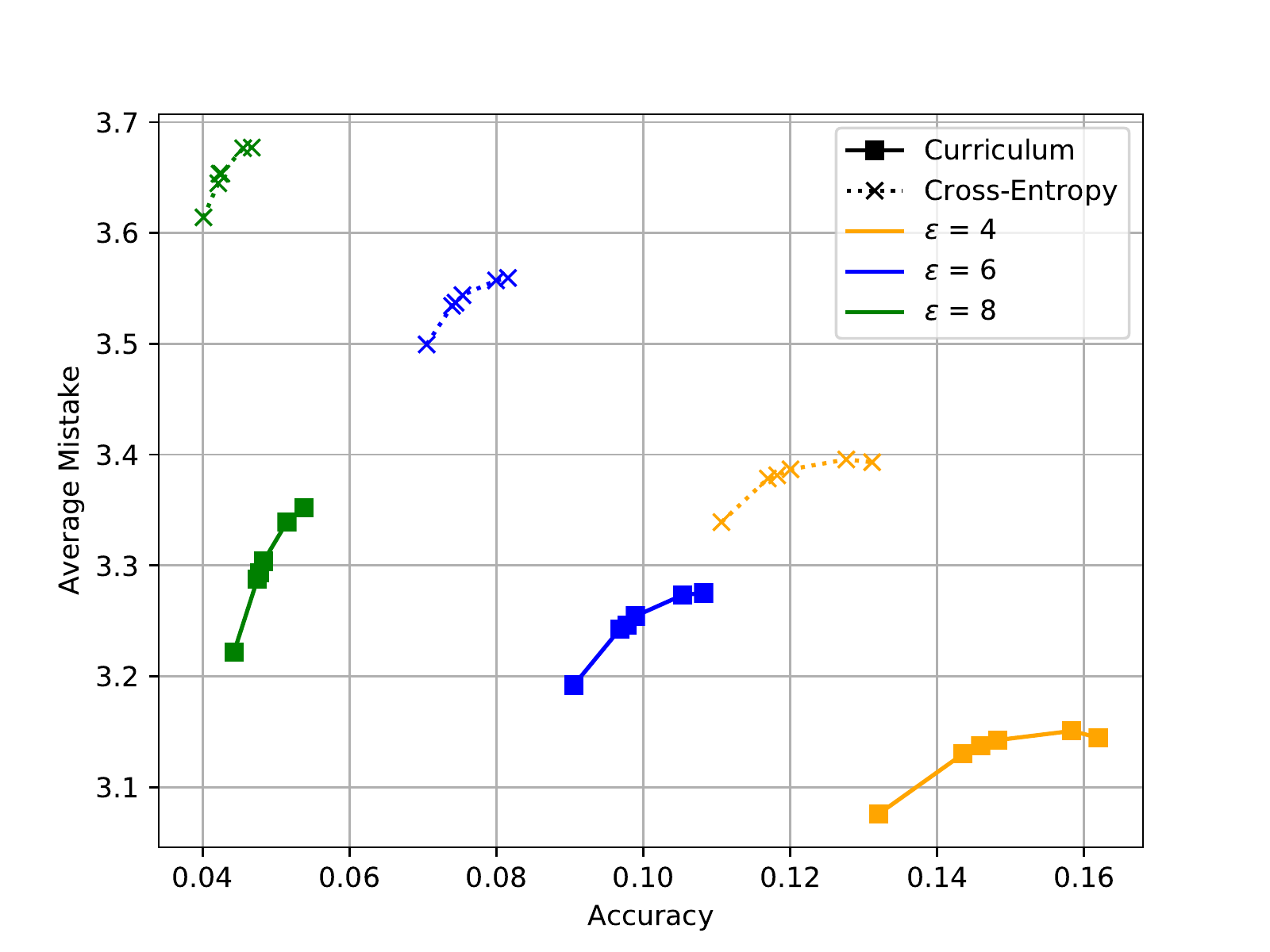}
        \caption{GHA Assessment}
     \end{subfigure}
     \hfill
     \begin{subfigure}[b]{0.33\textwidth}
        \includegraphics[width=\textwidth]{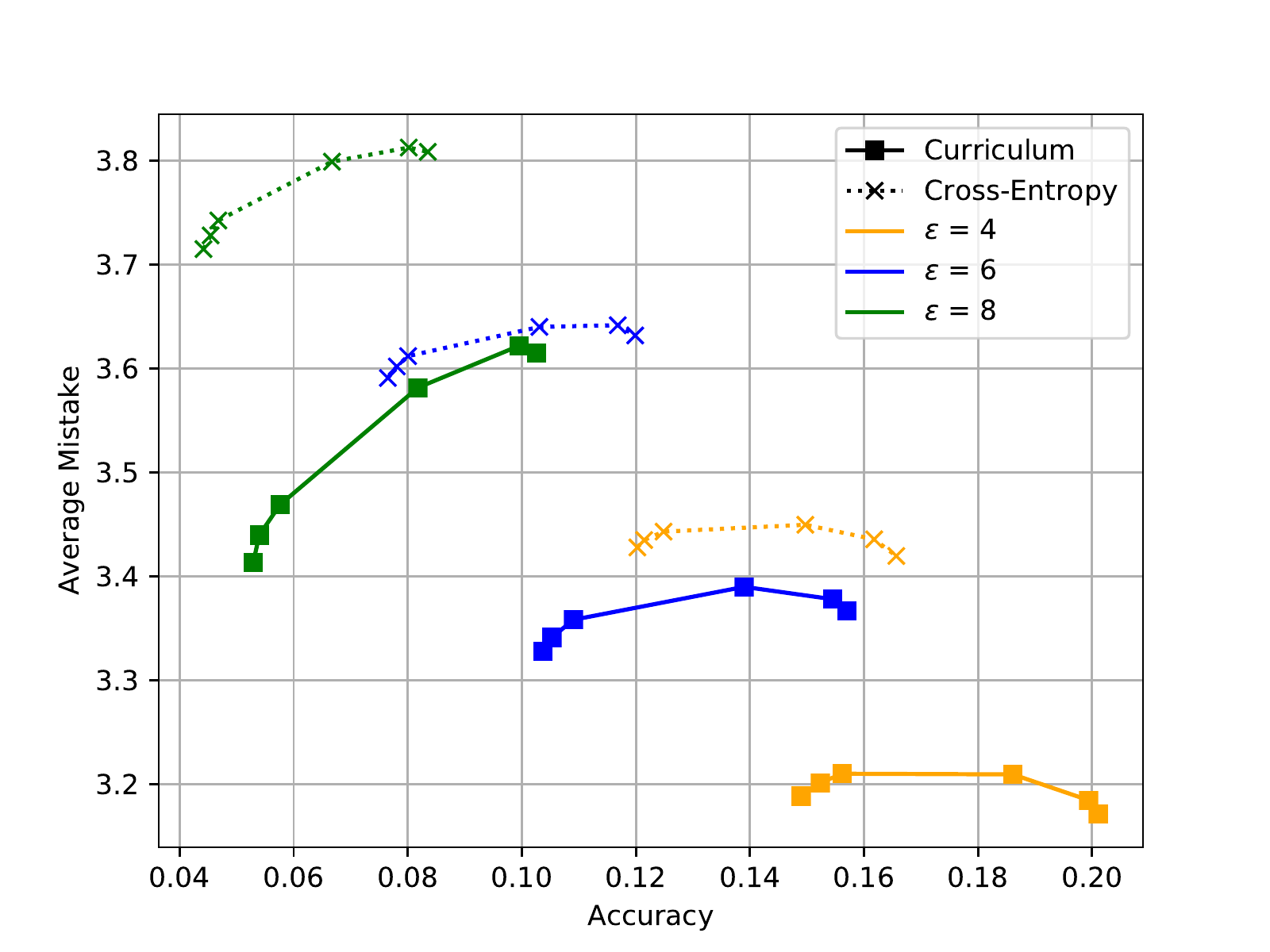}
        \caption{NHA Assessment}
     \end{subfigure}
     \caption{\textbf{TRADES Hierarchical Attack Evaluation.} We evaluate the performance of TRADES on our hierarchical attacks with 50 iterations at each target height on the validation set. The inclusion of our curriculum boost all metrics for all levels compared to the baseline. The results show similar behaviour than the results on Figure 3 of the manuscript.}
     \label{fig-appendix:trades_results}
     \vspace{-0.4cm}
 \end{figure*}

\section*{Hierarchical AutoAttack}

Our proposed attacks optimize the adversaries based on the probabilities of some chosen classes. So we adapted the AutoAttack benchmark to create the adversarial examples based on these chosen categories. We name this approach Hierarchical AutoAttack. For this experiment, we enhanced the AutoAttack with the NHA@$3$ attack. Furthermore, to boost the induced mistake, we harm all instances where the input image is correctly classified on the supernode at height $h$. We decided to choose the NHA attack as it provides the best Average Mistake increase. Also, we set $h=3$ since we consider that this setting provides the best trade-off between Accuracy and Adversarial Mistake. We chose to avoid experimentation over all the values of $h$ because AutoAttack is computational slow. We report the results of the Node-based Hierarchical AutoAttack (NHAA) on table \ref{tab-appendix:nha-aa}. The results show that CHAT is a defense mechanism that enjoys better protection against NHAA@$3$. 

\begin{table}[h]
    \centering
    \begin{tabular}{c|c|cc} \toprule
        $\boldsymbol{\epsilon}$ & \textbf{C} & \textbf{Acc}    & \textbf{AM}   \\ \midrule
        4 &            & 16.65  & 4.32 \\
        4 & \checkmark & 17.77  & 4.21 \\\midrule
        6 &            & 9.19   & 4.63 \\
        6 & \checkmark & 10.76  & 4.56 \\\midrule
        8 &            & 4.97   & 4.92 \\
        8 & \checkmark & 7.08   & 4.76 \\ \bottomrule
    \end{tabular}
    \caption{\textbf{Hierarchical AutoAttack.} We tested the AutoAttack enhanced with NHA@3. Similarly to all results on the main manuscript, our implementation enhances the robustness and reduce the severity of adversarial attacks.}
    \label{tab-appendix:nha-aa}
\end{table}

\section*{CHAT-enhanced TRADES}

For completeness, we train a model with TRADES and CHAT-enhanced TRADES (CHATeT). To achieve this CHATeT model, we replaced the second step of our curriculum with the traditional TRADES. Thus, it is necessary to minimizes the objective on Equation \ref{eq:trades}, where $L_{KL}$ is the Kullback-Leibler divergence, $L_{CE}$ the traditional cross entropy loss, and $\beta$ is a constant:

\begin{equation} \label{eq:trades}
    min \left\{L_{CE}(f(x), y) + \max_{x' \in B_{\ell_\infty}(x,\epsilon)} \beta L_{KL} (f(x), f(x')) \right\}
\end{equation}

\begin{table}[h]
    \centering
    \begin{tabular}{c|c|cc|cc} \toprule
      \multirow{2}{*}{$\boldsymbol{\epsilon}$} & \multirow{2}{*}{\textbf{C}} & \multicolumn{2}{c|}{\textbf{Clean}} & \multicolumn{2}{c}{\textbf{PGD}} \\
          &            & \textbf{Acc}   & \textbf{AM}   & \textbf{Acc}   & \textbf{AM}   \\ \midrule
        4 &            & 23.90 & 3.33 & 11.06 & 3.34 \\
        4 & \checkmark & 29.23 & 3.03 & 13.20 & 3.08 \\ \midrule
        6 &            & 21.26 & 3.49 & 7.04  & 3.50 \\
        6 & \checkmark & 27.91 & 3.14 & 9.05  & 3.19 \\ \midrule
        8 &            & 19.94 & 3.59 & 4.01  & 3.61 \\
        8 & \checkmark & 29.84 & 3.11 & 4.47  & 3.22 \\ \bottomrule
    \end{tabular}
    \caption{\textbf{Effect of CHAT with TRADES.} Enhancing TRADES with our proposed mechanism boost the accuracy performance and decreases the average mistake on both clean and adversarial settings.}
    \label{tab-appendix:trades-pgd}
\end{table}

For $\epsilon \in \{\nicefrac{6}{255}, \nicefrac{8}{255}\}$, we trained the standard and CHAT-enhanced models with a step size $\alpha$ of $\nicefrac{3}{255}$ and $\nicefrac{4}{255}$ for $3$ or $4$ iterations $I$. Then we choose the best performing model, which are $(\alpha, I) = (\nicefrac{3}{255}, 3)$ for TRADES and $(\nicefrac{4}{255}, 4)$ for CHATeT. For $\epsilon = \nicefrac{4}{255}$, we used $\alpha=1$ with $I=4$ for the vanilla and enhanced model. Further, we set the constant $\beta = 6$ for all models. We pick the value of $\beta$ through a hyperparameter search with a perturbation budget of $\epsilon = \nicefrac{4}{255}$. On table \ref{tab-appendix:trades-pgd} we display the results of the experiments with CHATeT with PGD50 and on Figure \ref{fig-appendix:trades_results} the results against the hierarchical-aware attacks, both on the validation set. The results show that CHAT does not solely boost the performance for FAT but also TRADES. To our surprise, the CHATeT's clean accuracy was higher among all models. We suspect that our coarse hyperparameter search did not yield the best performance. Nonetheless, we achieve a performance gain on both metrics.

\section*{Adversarial Images}

\begin{figure*}
    \centering
    \includegraphics[width=\textwidth]{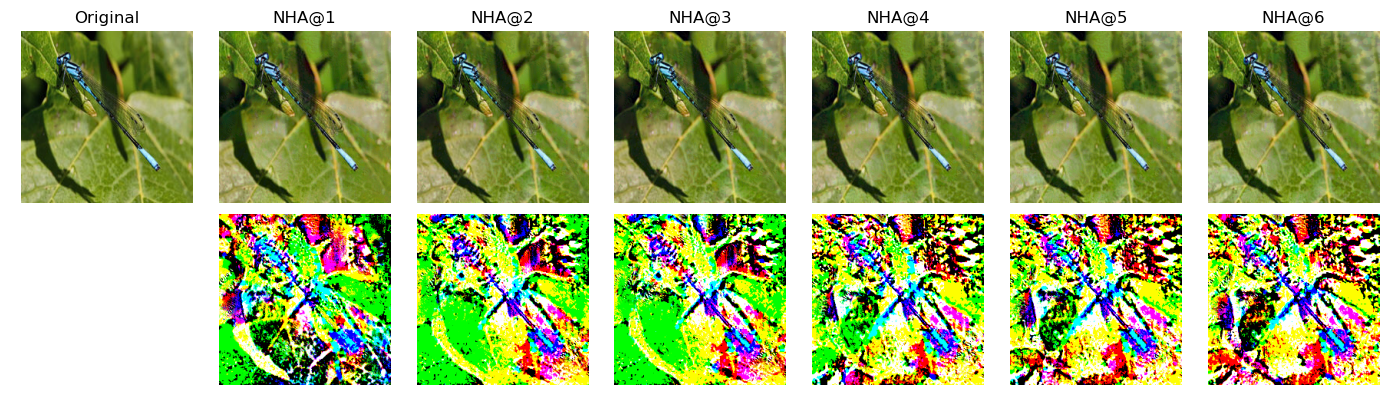}
    \caption{NHA Adversarial examples}
    \label{fig-a:first-nha}
\end{figure*}

\begin{figure*}
    \centering
    \includegraphics[width=\textwidth]{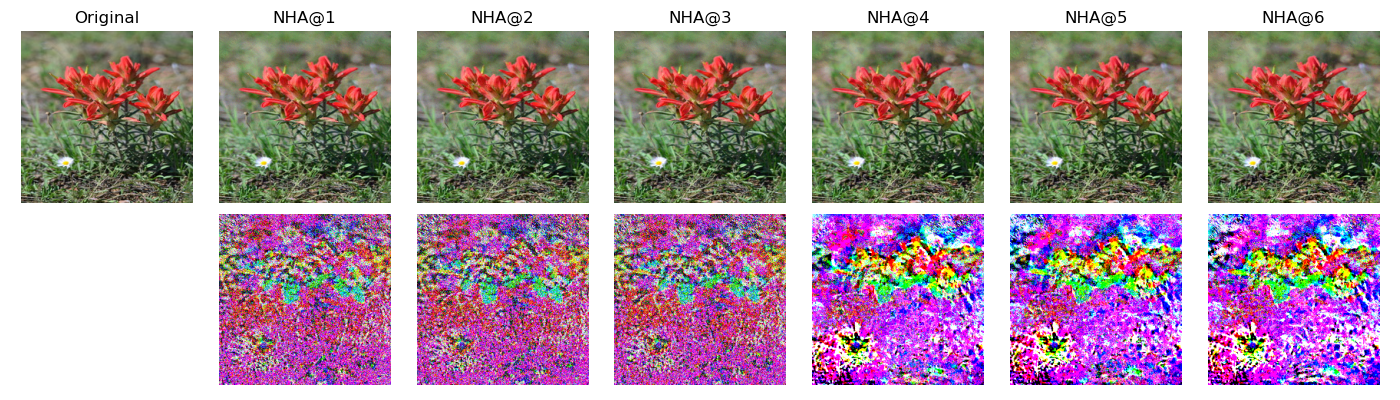}
    \caption{NHA Adversarial examples}
    \label{fig-a:last-nha}
\end{figure*}

\begin{figure*}
    \centering
    \includegraphics[width=\textwidth]{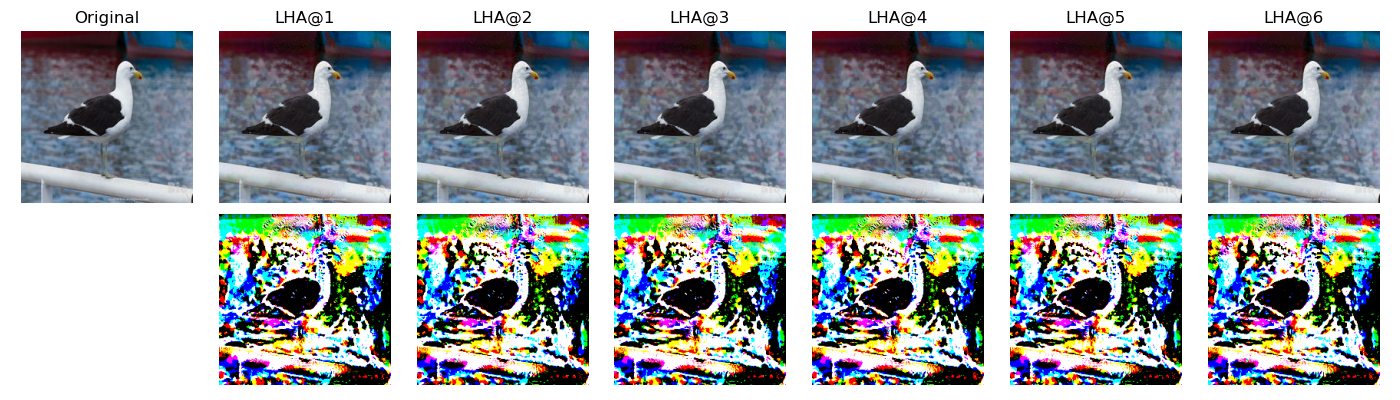}
    \caption{GHA Adversarial examples}
    \label{fig-a:first-gha}
\end{figure*}

\begin{figure*}
    \centering
    \includegraphics[width=\textwidth]{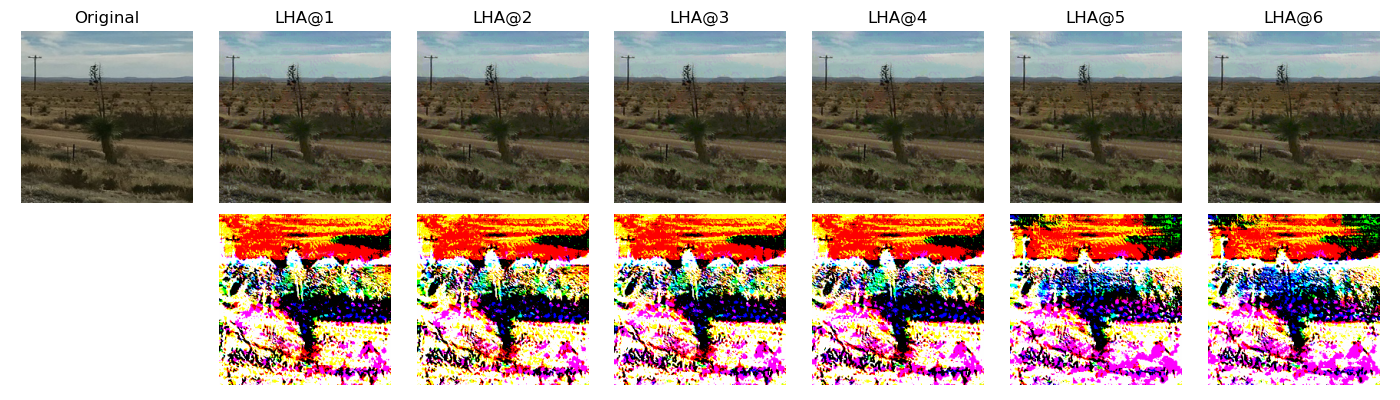}
    \caption{GHA Adversarial examples}
    \label{fig-a:last-gha}
\end{figure*}

\begin{figure*}
    \centering
    \includegraphics[width=\textwidth]{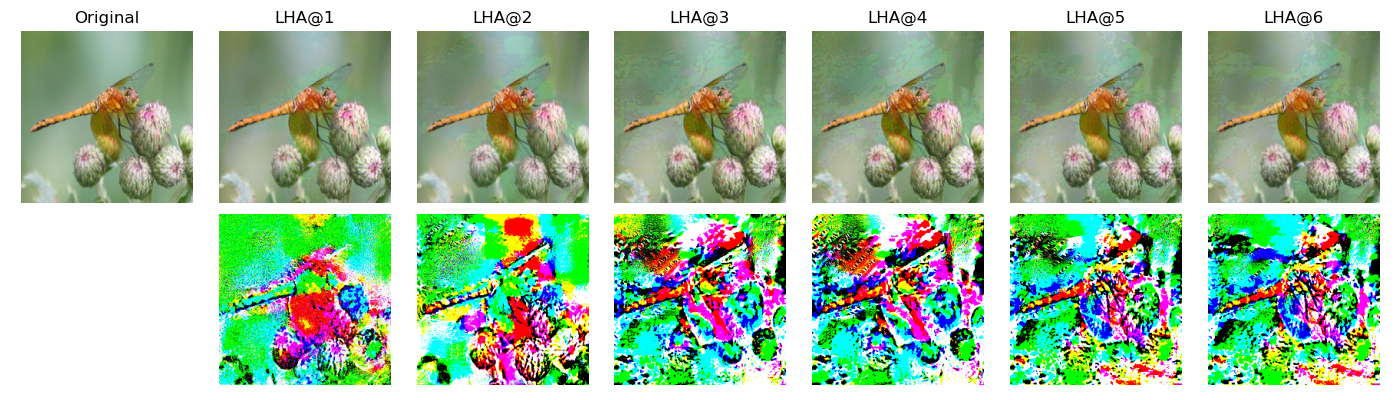}
    \caption{LHA Adversarial examples}
    \label{fig-a:first-lha}
\end{figure*}

\begin{figure*}
    \centering
    \includegraphics[width=\textwidth]{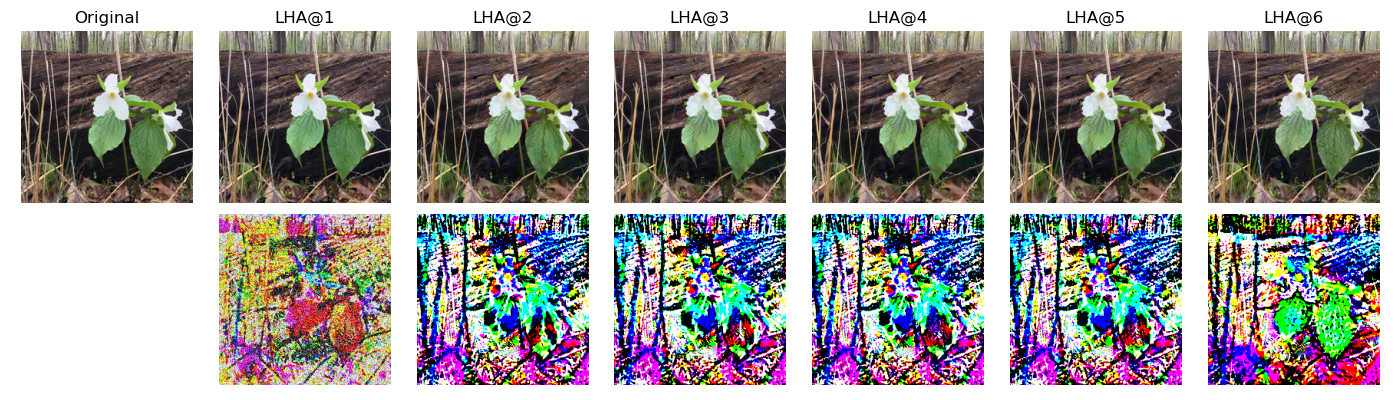}
    \caption{LHA Adversarial examples}
    \label{fig-a:last-lha}
\end{figure*}

We present in table \ref{tab-appendix:main_results} the results on the validation set of the hierarchical attacks against FAT and CHAT. Furthermore, we display some adversarial images created by all our attacks. We show the original image, the adversary, and the noise. We only display the adversaries of the best model for $\epsilon=8$ to visualize the perturbation on the image. Figures \ref{fig-a:first-nha} to \ref{fig-a:last-nha} display adversaries generated by NHA. The adversaries for GHA are between Figure \ref{fig-a:first-gha} and \ref{fig-a:last-gha}. Lastly, Figures \ref{fig-a:first-lha} to \ref{fig-a:last-lha} show some LHA adversaries.

\begin{table*}[t]
    \centering
    \small
    \begin{tabular}{c|c|c||cc|cc|cc|cc|cc|cc} \cmidrule[\heavyrulewidth]{2-15}
         & \multicolumn{2}{c||}{Level} & \multicolumn{2}{c|}{1} & \multicolumn{2}{c|}{2} & \multicolumn{2}{c|}{3} & \multicolumn{2}{c|}{4} & \multicolumn{2}{c|}{5} & \multicolumn{2}{c}{6} \\ \cmidrule{2-15}
        &$\epsilon$& C  & Acc   & AM    & Acc   & AM    & Acc   & AM    & Acc   & AM    & Acc   & AM    & Acc   & AM    \\ \midrule[\heavyrulewidth]  \midrule[\heavyrulewidth] 
        \multirow{6}{*}{\rotatebox[origin=c]{90}{LHA}}&4 &            & 14.64 &  2.98 & 14.39 & 2.99  & 13.80 & 3.02  & 12.39 & 3.12  & 12.23 & 3.15  & 12.25 &  3.17 \\
        &4 & \checkmark & 16.32 &  2.81 & 16.01 & 2.82  & 15.36 & 2.86  & 13.60 & 2.97  & 13.36 &  3.01 & 13.33 &  3.02 \\ \cmidrule{2-15}
        &6 &            & 8.99  & 3.23  & 8.78  & 3.23  & 8.33  & 3.27  & 7.29  & 3.36  & 7.09  & 3.39  & 7.08  & 3.41  \\
        &6 & \checkmark & 10.71 & 3.06  & 10.49 & 3.07  & 9.92  & 3.10  & 8.53  & 3.20  & 8.36  & 3.24  & 8.32  & 3.26  \\ \cmidrule{2-15}
        &8 &            & 5.54  & 3.53  & 5.45  & 3.54  & 5.13  & 3.57  & 4.31  & 3.64  & 4.24  & 3.67  & 4.27  & 3.69  \\
        &8 & \checkmark & 7.65  & 3.27  & 7.52  & 3.28  & 7.05  & 3.32  & 6.11  & 3.41  & 5.92  & 3.45  & 5.98  & 3.46  \\ \midrule[\heavyrulewidth] \midrule[\heavyrulewidth] 
        \multirow{6}{*}{\rotatebox[origin=c]{90}{GHA}}&4 &            & 12.33 &  3.20 & 14.71 & 3.32  & 15.07 & 3.33  & 15.56 &  3.34 & 17.52 &  3.34 & 18.42 & 3.33 \\
        &4 & \checkmark & 13.36 &  3.05 & 15.97 & 3.18  & 16.47 & 3.19  & 17.01 &  3.20 & 19.33 &  3.21 & 20.39 & 3.19 \\ \cmidrule{2-15}
        &6 &            & 7.13  & 3.44  & 8.55  & 3.54  & 8.82  & 3.55  & 9.14  & 3.56  & 10.69 & 3.58  & 11.60 & 3.58  \\
        &6 & \checkmark & 8.34  & 3.29  & 10.20 & 3.40  & 10.55 & 3.41  & 10.89 & 3.43  & 12.75 & 3.44  & 13.79 & 3.44  \\ \cmidrule{2-15}
        &8 &            & 4.31  & 3.71  & 5.21  & 3.79  & 5.35  & 3.79  & 5.53  & 3.80  & 6.60  & 3.82  & 7.32  & 3.82  \\
        &8 & \checkmark & 6.05  & 3.49  & 7.28  & 3.58  & 7.50  & 3.58  & 7.81  & 3.60  & 9.26  & 3.62  & 10.10 & 3.62  \\ \midrule[\heavyrulewidth] \midrule[\heavyrulewidth] 
        \multirow{6}{*}{\rotatebox[origin=c]{90}{NHA}}&4 &            & 15.12 & 3.38  & 15.59 & 3.39  & 16.39 &  3.41 & 20.56 & 3.39  & 22.54 &  3.39 & 23.12 &  3.36 \\
        &4 & \checkmark & 16.28 & 3.23  & 16.87 & 3.25  & 17.58 &  3.27 & 22.03 & 3.27  & 23.92 & 3.24  & 24.46 & 3.22 \\ \cmidrule{2-15}
        &6 &            & 8.63  & 3.59  & 8.88  & 3.60  & 9.29  & 3.62  & 12.85 & 3.66  & 14.71 & 3.65  & 15.24 & 3.63  \\
        &6 & \checkmark & 10.29 & 3.45  & 10.54 & 3.47  & 11.09 & 3.49  & 15.20 & 3.52  & 16.96 & 3.51  & 17.49 & 3.49  \\ \cmidrule{2-15}
        &8 &            & 5.27  & 3.84  & 5.39  & 3.85  & 5.65  & 3.87  & 8.06  & 3.92  & 9.93  & 3.91  & 10.27 & 3.90  \\
        &8 & \checkmark & 7.21  & 3.63  & 7.49  & 3.64  & 7.84  & 3.66  & 11.00 & 3.71  & 12.92 & 3.70  & 13.46 & 3.69  \\ \bottomrule
        
    \end{tabular}
    \caption{\textbf{Hierarchical Attack Evaluation.} We evaluate the performance of FAT on our hierarchical attacks with 50 iterations on each level. The inclusion of our curriculum boost all metrics for all levels compared to the baseline. Acc stands for Accuracy and AM Average Mistake. LHA is the strongest accuracy-related attack among the proposed ones. It even surpasses the PGD when setting the level $l$ to 5 or 6. The GHA enjoys a balanced between severity and accuracy. The NHA is the most severe but less successful attack among the set of hierarchical attacks.}
    \label{tab-appendix:main_results}
\end{table*}


%

\begin{figure*}
    \centering
    \includegraphics[width=\textwidth]{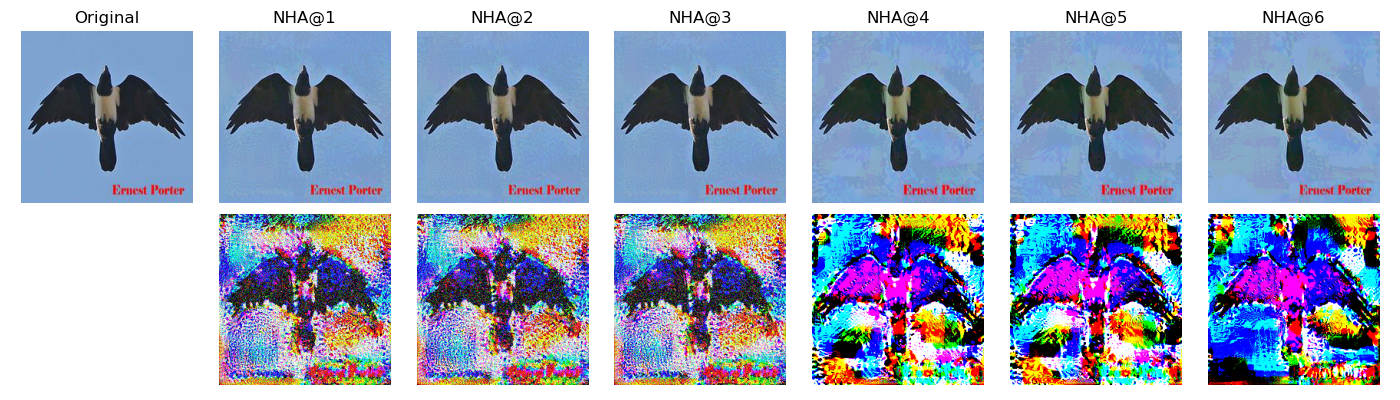}
    \caption{NHA Adversarial examples}
\end{figure*}


\begin{figure*}
    \centering
    \includegraphics[width=\textwidth]{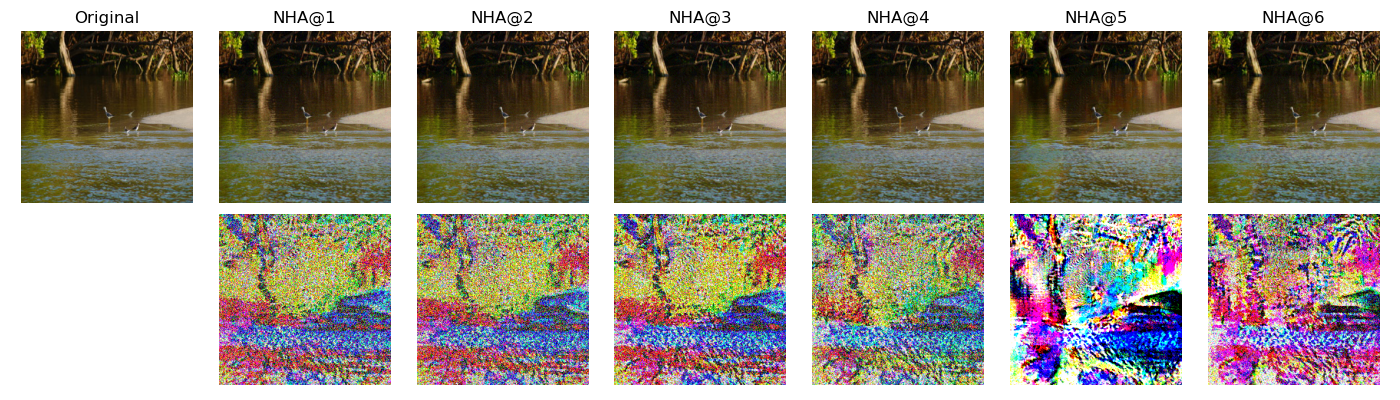}
    \caption{NHA Adversarial examples}
\end{figure*}


\begin{figure*}
    \centering
    \includegraphics[width=\textwidth]{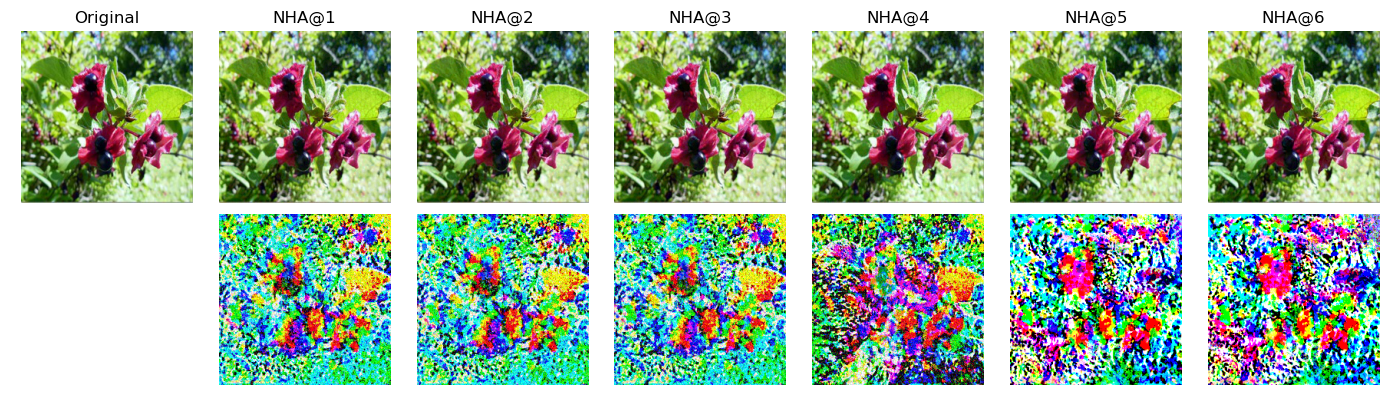}
    \caption{NHA Adversarial examples}
\end{figure*}

\begin{figure*}
    \centering
    \includegraphics[width=\textwidth]{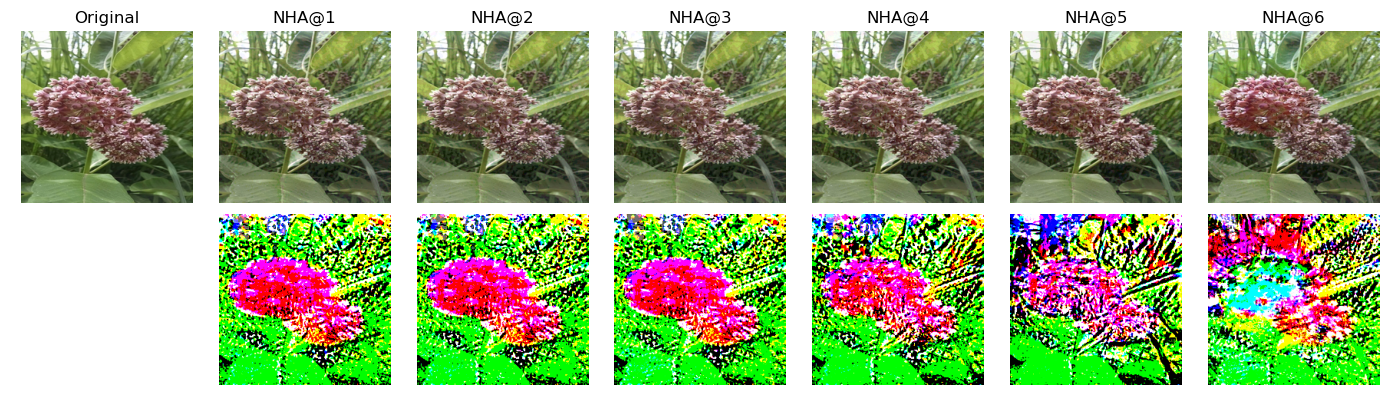}
    \caption{NHA Adversarial examples}
\end{figure*}

\begin{figure*}
    \centering
    \includegraphics[width=\textwidth]{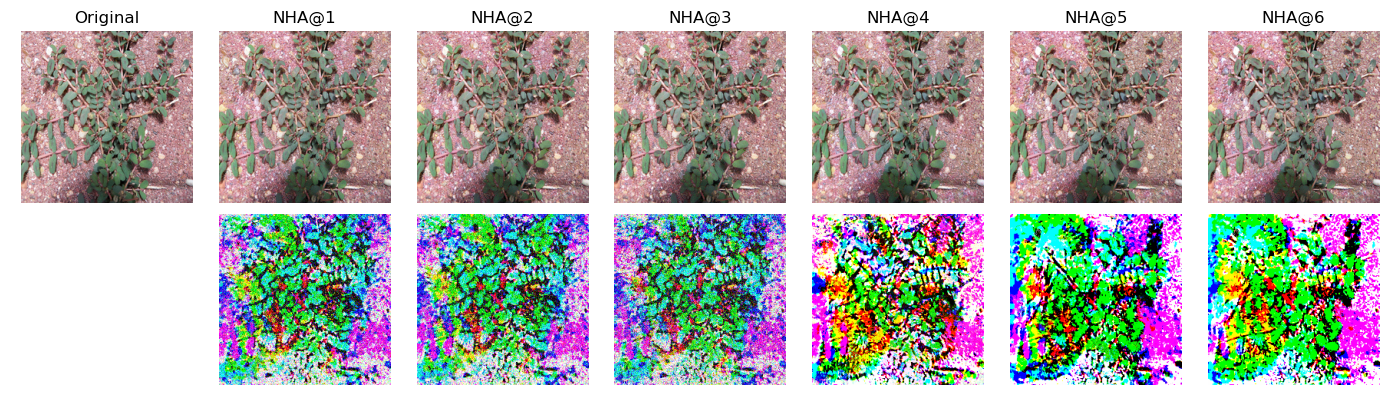}
    \caption{NHA Adversarial examples}
\end{figure*}

\begin{figure*}
    \centering
    \includegraphics[width=\textwidth]{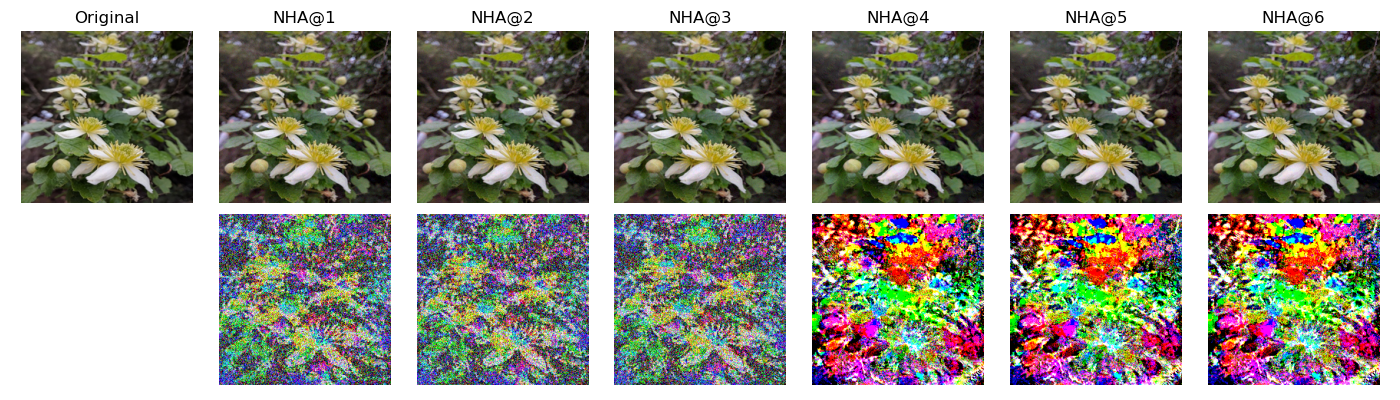}
    \caption{NHA Adversarial examples}
\end{figure*}

%

\begin{figure*}
    \centering
    \includegraphics[width=\textwidth]{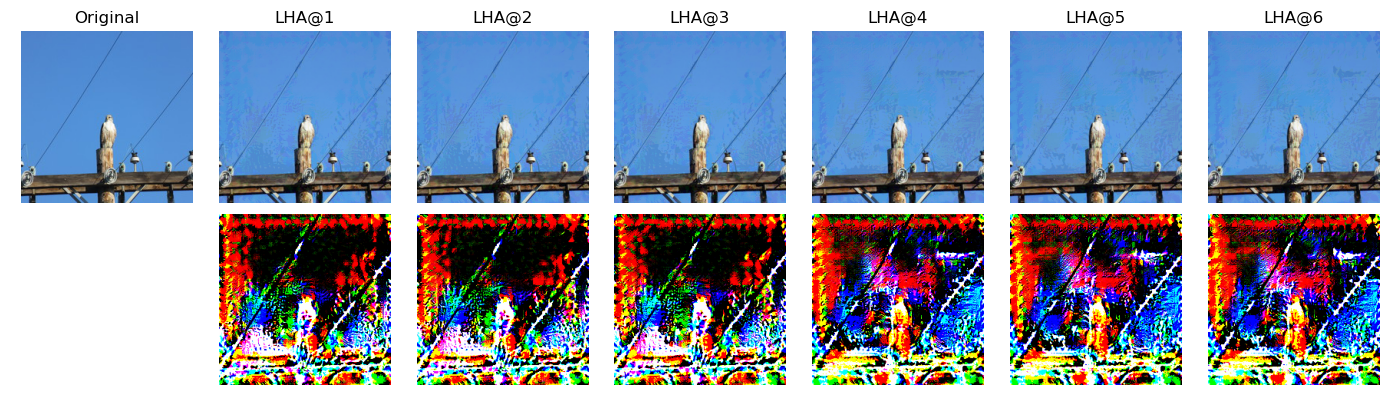}
    \caption{GHA Adversarial examples}
\end{figure*}


\begin{figure*}
    \centering
    \includegraphics[width=\textwidth]{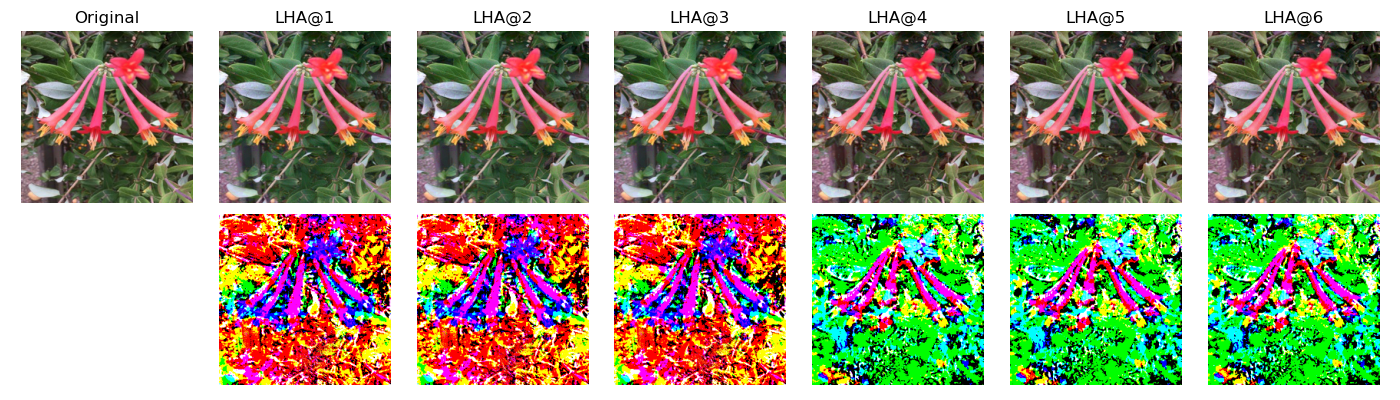}
    \caption{GHA Adversarial examples}
\end{figure*}

\begin{figure*}
    \centering
    \includegraphics[width=\textwidth]{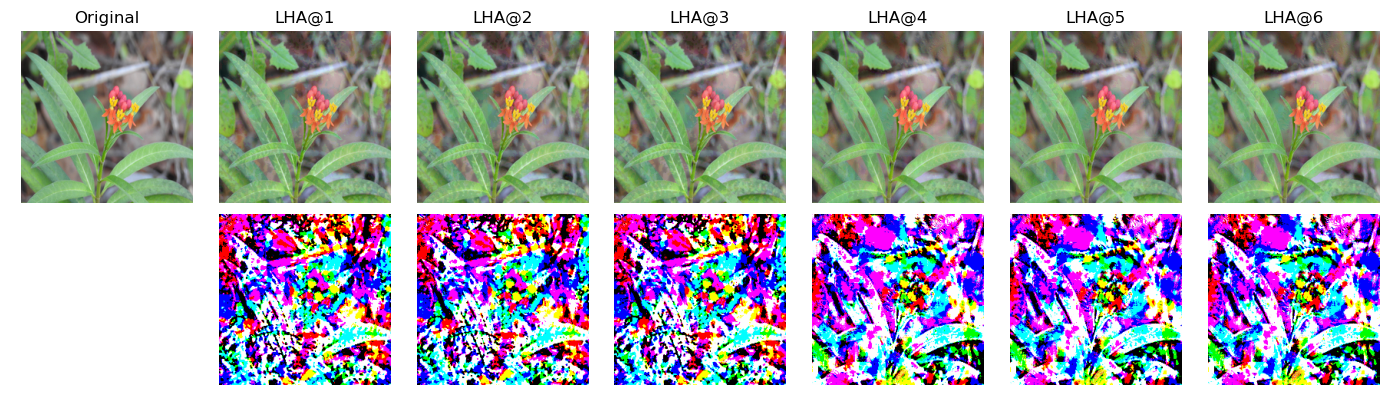}
    \caption{GHA Adversarial examples}
\end{figure*}

\begin{figure*}
    \centering
    \includegraphics[width=\textwidth]{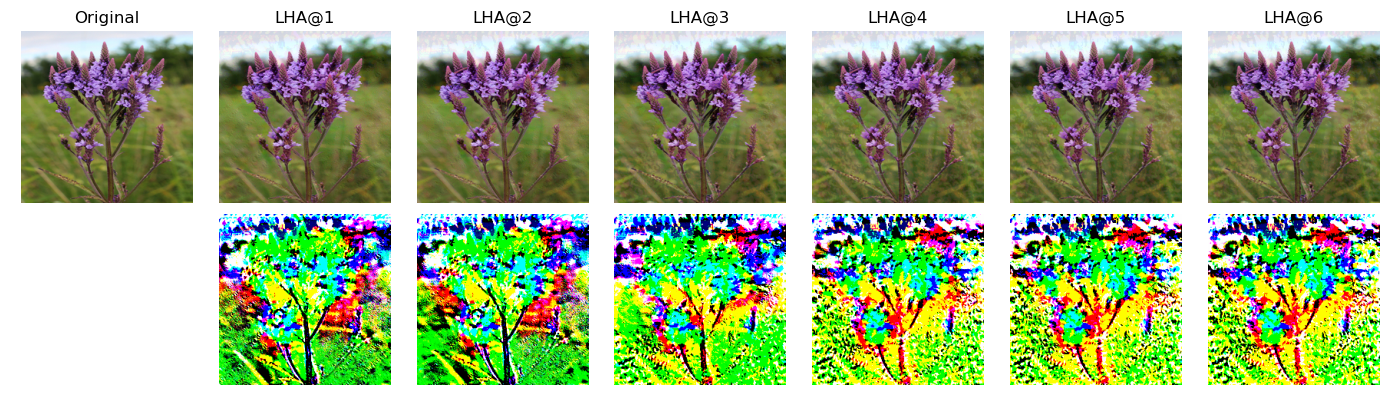}
    \caption{GHA Adversarial examples}
\end{figure*}


\begin{figure*}
    \centering
    \includegraphics[width=\textwidth]{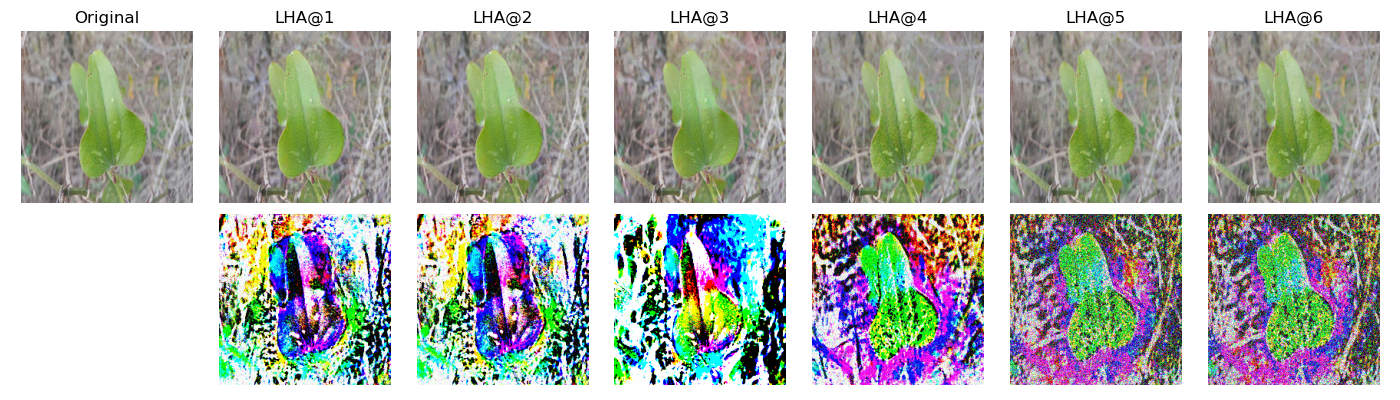}
    \caption{GHA Adversarial examples}
\end{figure*}

\begin{figure*}
    \centering
    \includegraphics[width=\textwidth]{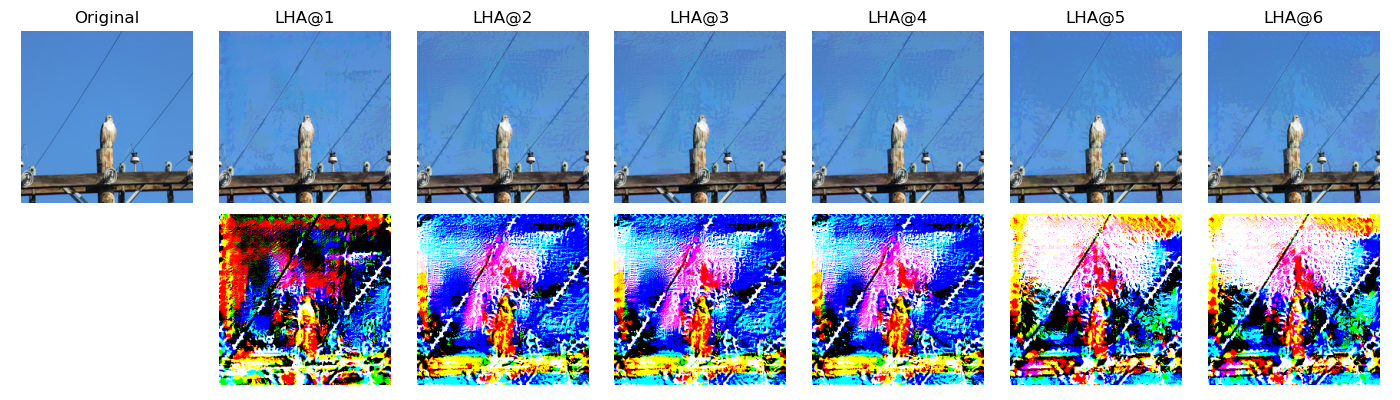}
    \caption{LHA Adversarial examples}
\end{figure*}

\begin{figure*}
    \centering
    \includegraphics[width=\textwidth]{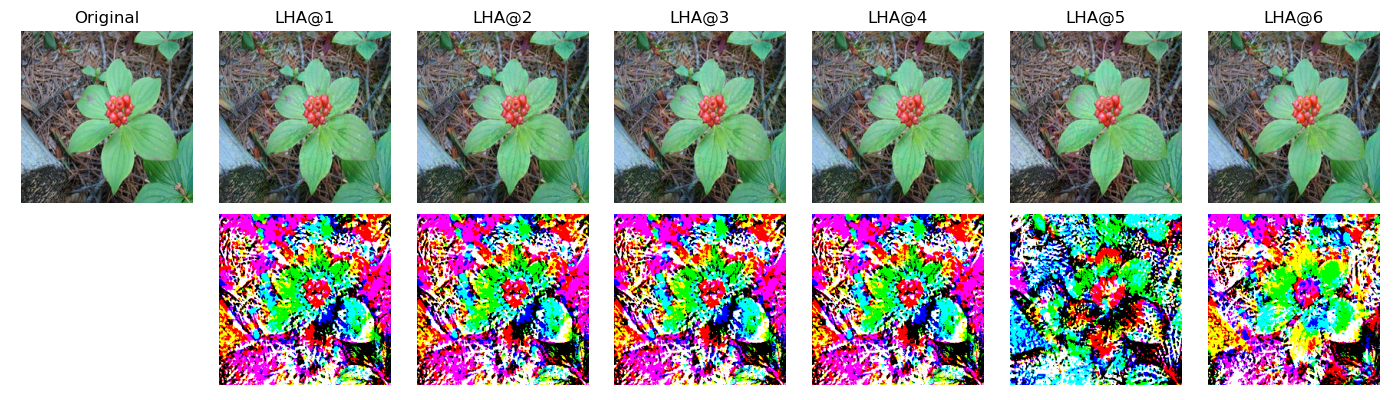}
    \caption{LHA Adversarial examples}
\end{figure*}


\begin{figure*}
    \centering
    \includegraphics[width=\textwidth]{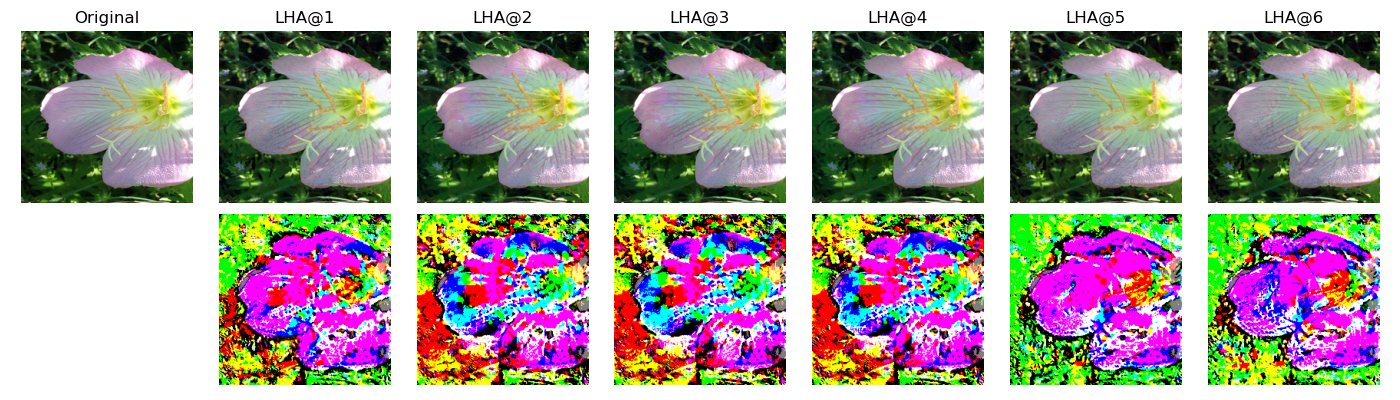}
    \caption{LHA Adversarial examples}
\end{figure*}

\begin{figure*}
    \centering
    \includegraphics[width=\textwidth]{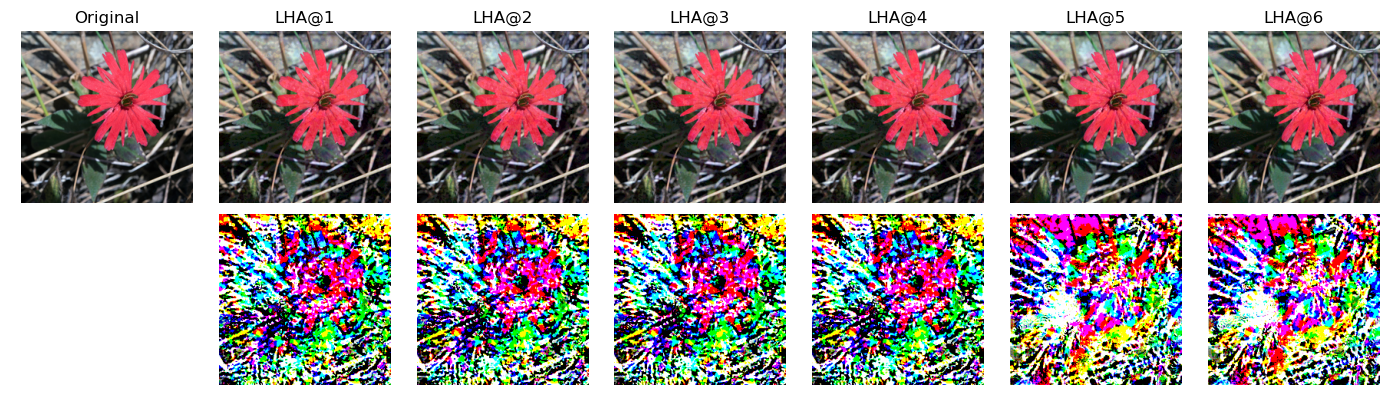}
    \caption{LHA Adversarial examples}
\end{figure*}

\begin{figure*}
    \centering
    \includegraphics[width=\textwidth]{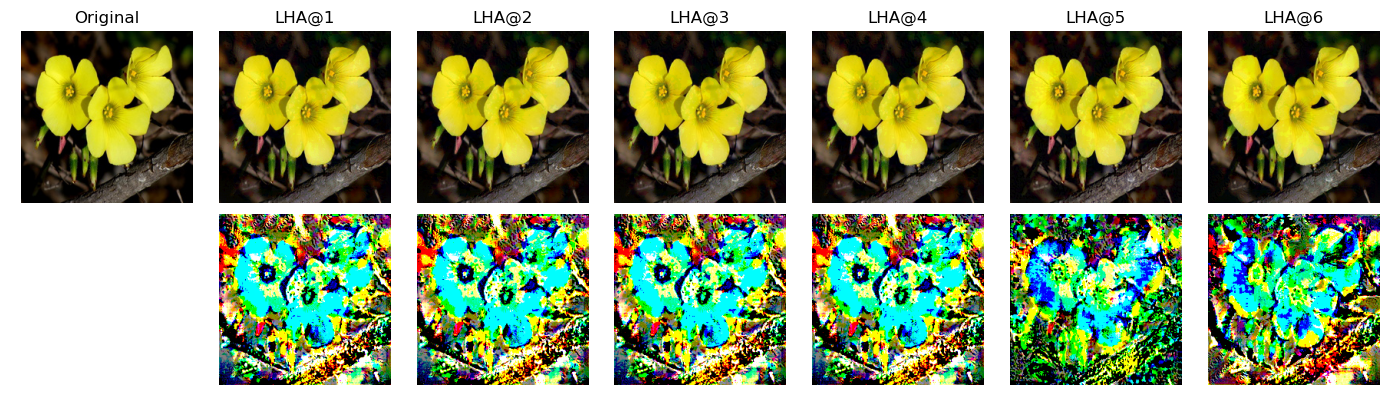}
    \caption{LHA Adversarial examples}
\end{figure*}



\end{document}